\documentclass[final]{cvpr}
\usepackage{color}
\usepackage{array}
\usepackage{times}
\usepackage{epsfig}
\usepackage{graphicx}
\usepackage{amsmath}
\usepackage{amssymb}
\usepackage{lipsum}
\usepackage{subfiles}
\usepackage{subcaption}
\usepackage{tikz}
\usepackage{pgfplots}
\pgfplotsset{compat=1.16}
\usepackage{enumitem}
\usepackage{booktabs}
\usepackage[colorlinks, linkcolor=red]{hyperref}
\definecolor{red}{rgb}{1,0,0}
\definecolor{blue}{rgb}{1,0.843,0}
\usepackage[accsupp]{axessibility} 
\usepackage{algorithm}
\usepackage{algpseudocode}

\newcolumntype{x}[1]{>{\centering\arraybackslash}p{#1pt}}
\newlength\savewidth
\newcommand{\tablestyle}[2]{\setlength{\tabcolsep}{#1}\renewcommand{\arraystretch}{#2}\centering\footnotesize}

\def\cvprPaperID{2121} 
\def\confYear{CVPR 2022}

\definecolor{tp}{HTML}{3BA3EC}
\definecolor{fp}{HTML}{FF6100}

\colorlet{mr}{tp}
\colorlet{ji}{orange!60}
\colorlet{ap}{red!70}

\author{\Large 
Anlin Zheng\textsuperscript{\rm 1,3*},
Yuang Zhang\textsuperscript{\rm 2*},
Xiangyu Zhang\textsuperscript{\rm 1},
Xiaojuan Qi\textsuperscript{\rm 3},
Jian Sun\textsuperscript{\rm 1}\\
\textsuperscript{\rm 1} MEGVII Technology \quad
\textsuperscript{\rm 2}Shanghai Jiao Tong University\quad
\textsuperscript{\rm 3}{University of Hong Kong}
\\
\texttt{\small zyayoung@sjtu.edu.cn, xjqi@eee.hku.hk, \{zhenganlin, zhangxiangyu, sunjian\}@megvii.com}\\
}

\begin{document}

\title{Progressive End-to-End Object Detection in Crowded Scenes}


\maketitle

{\let\thefootnote\relax\footnotetext{* Equal contribution.}}

\begin{abstract}
In this paper, we propose a new query-based detection framework for crowd detection. Previous query-based detectors suffer from two drawbacks: first, multiple predictions will be inferred for a single object, typically in crowded scenes; second, the performance saturates as the depth of the decoding stage increases. Benefiting from the nature of the one-to-one label assignment rule, we propose a progressive predicting method to address the above issues. Specifically, we first select accepted queries prone to generate true positive predictions, then refine the rest noisy queries according to the previously accepted predictions. Experiments show that our method can significantly boost the performance of query-based detectors in crowded scenes. Equipped with our approach, Sparse RCNN achieves 92.0\% $\text{AP}$, 41.4\% $\text{MR}^{-2}$ and 83.2\% $\text{JI}$ on the challenging CrowdHuman~\cite{shao2018crowdhuman} dataset, outperforming the box-based method MIP~\cite{chu2020detection} that specifies in handling crowded scenarios. Moreover, the proposed method, robust to crowdedness, can still obtain consistent improvements on moderately and slightly crowded datasets like CityPersons~\cite{zhang2017citypersons} and COCO~\cite{lin2014microsoft}. Code will be made publicly available at https://github.com/megvii-model/Iter-E2EDET.

\vspace{-2pt}
\end{abstract}

\section{Introduction}\label{intro}
Crowded object detection is a practical yet challenging research field in computer vision. Many research efforts have been made and achieved impressive progress ~\cite{lu2019semantic, chi2020pedhunter, chi2020relational,zhang2019double,zhang2018occlusion,chu2020detection,iterdet2021,psrcnn,lin2020detr} in the last few decades. However, most of them~\cite{lu2019semantic, chi2020pedhunter, chi2020relational,zhang2019double,zhang2018occlusion,chu2020detection,iterdet2021,psrcnn} require hand-craft components, e.g.\ anchor settings and post-processing, resulted in sub-optimal performance in handling scenes.

\begin{figure}[!t]
\begin{center}
 \includegraphics[width=1.\linewidth]{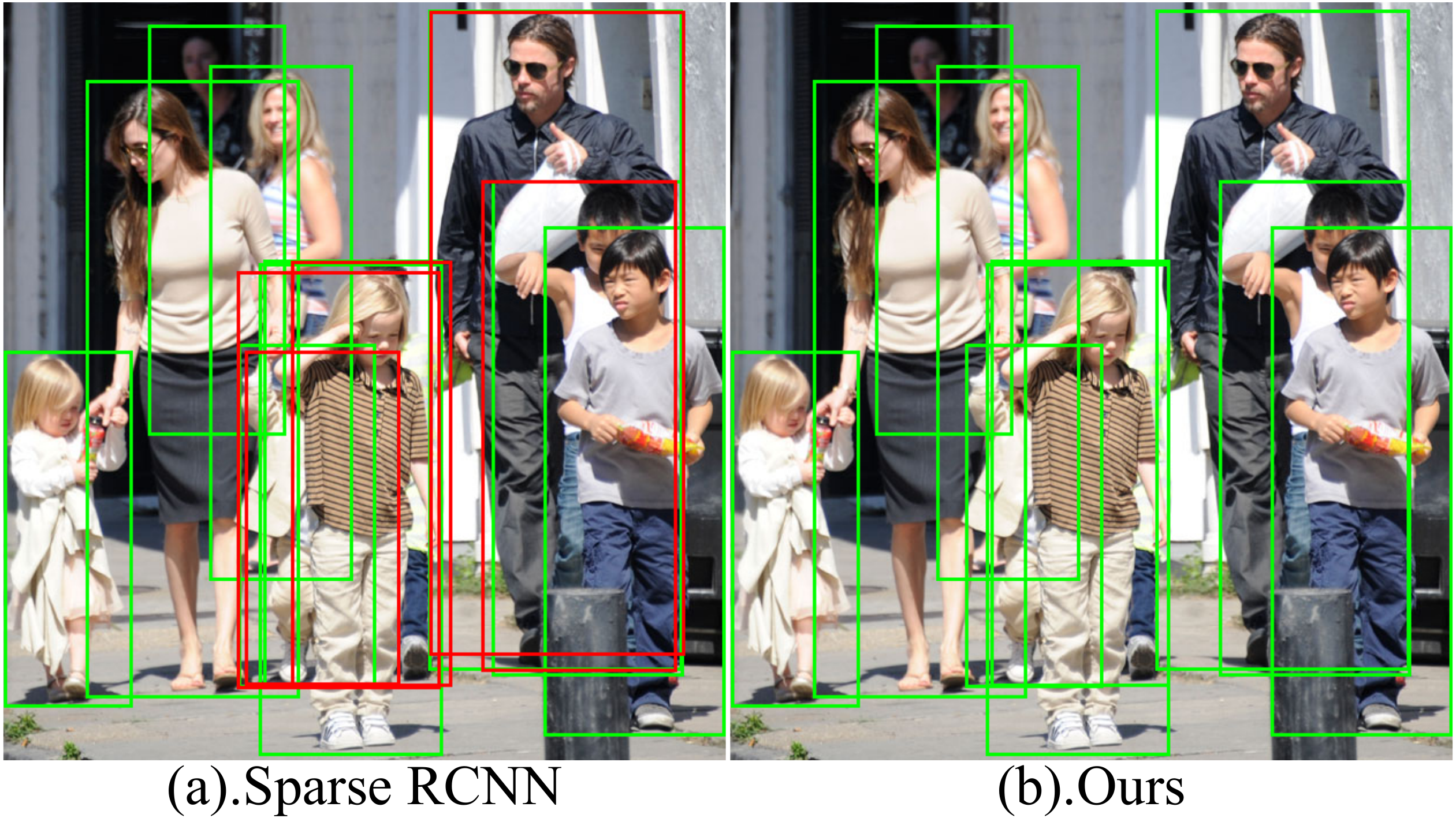}
\vspace{-7ex}
\end{center}
   \caption{\textcolor{red}{1a}. Sparse RCNN~\cite{sun2020sparse} introduces \textit{false positives} in crowded scenes. \textcolor{red}{1b}. Our approach can remove those \textit{false positives} and ensure each object can be detected only once. Green boxes indicate \emph{true positives} while red ones represent \emph{false positives}.}
\label{fig:visulization}
\vspace{-2ex}
\end{figure}

\begin{figure}
  \footnotesize
  \centering
  \begin{subfigure}{0.49\linewidth}
    \centering
    \definecolor{tp}{HTML}{3BA3EC}
\definecolor{fp}{HTML}{FF6100}

\hspace{-1em}
\begin{tikzpicture}
  \scriptsize
  \begin{axis}[
      xlabel={score},
      ylabel=ratio of ratio,
      ybar, axis on top,
      height=4.5cm, width=1.2\linewidth,
      bar width=0.18cm,
      ybar=0,
      ymajorgrids,
      tick align=inside,
      enlarge y limits={value=.1,upper},
      enlarge x limits=.1,
      ymin=0, ymax=0.5,
      axis x line*=bottom,
      axis y line*=left,
      y axis line style={opacity=0},
      major tick length=2pt,
      ytick style={draw=none},
      legend style={
          draw=none,
          fill=none,
          at={(0.5, 0.675)},
          inner sep=1.2,
          anchor=north,
          legend columns=-1,
          /tikz/every even column/.append style={column sep=0.1cm}
        },
      legend image code/.code={\draw [#1, draw=none] (0cm,-0.05cm) rectangle (0.18cm,0.08cm); },
      x label style={at={(axis description cs:.5,-0.004cm)},anchor=north},
      ylabel near ticks,
      xtick={0.2, 0.3, 0.4, 0.5, 0.6, 0.7, 0.8, 0.9, 1.0},
      xticklabels={.2, .3, .4, .5, .6, .7, .8, .9, 1},
      ytick={.00, .05, .10, .15, .20, .25, .30},
      yticklabels={.00, .05, .10, .15, .20, .25, .30},
    ]
    \addplot [draw=none, fill=tp] coordinates {
        (0.25, 0.0364746622584506)
        (0.35, 0.0435073818617112)
        (0.45, 0.0463361454711280)
        (0.55, 0.0477042125064128)
        (0.65, 0.0497135609644872)
        (0.75, 0.0531479792509833)
        (0.85, 0.0740395029356438)
        (0.95, 0.2959656273157384) };
    \addplot [draw=none,fill=fp] coordinates {
        (0.25, 0.2390412130194379)
        (0.35, 0.0692370176138630)
        (0.45, 0.0240480533546143)
        (0.55, 0.0103745083509091)
        (0.65, 0.0052300062703072)
        (0.75, 0.0025508749928746)
        (0.85, 0.0013466909878584)
        (0.95, 0.0012825628455794) };

    \legend{\emph{TP}, \emph{FP}}
  \end{axis}
  \begin{axis}[
      ybar, axis on top,
      height=4.5cm, width=1.2\linewidth,
      bar width=0.18cm,
      ybar=0,
      ymajorgrids,
      tick align=inside,
      enlarge y limits={value=.1,upper},
      enlarge x limits=.1,
      ymin=-0.3, ymax=0.03,
      axis x line*=top,
      axis y line*=left,
      xlabel={absolute relative improvement},
      x axis line style={opacity=0},
      y axis line style={opacity=0},
      major tick length=2pt,
      xtick style={draw=none},
      ytick style={draw=none},
      x label style={at={(axis description cs:.5,.9)},anchor=south},
      ylabel near ticks,
      ytick={-.03, .00, .03},
      yticklabels={-.03, .00, .03},
      xtick={},
      xticklabels={},
    ]
    \addplot [draw=none, fill=tp] coordinates {
        (0.25, 0.035896917005-0.0364746622584506)
        (0.35, 0.041142910514-0.0435073818617112)
        (0.45, 0.045275881661-0.0463361454711280)
        (0.55, 0.049214268479-0.0477042125064128)
        (0.65, 0.046256586679-0.0497135609644872)
        (0.75, 0.075872321546-0.0531479792509833)
        (0.85, 0.090396095860-0.0740395029356438)
        (0.95, 0.326738221810-0.2959656273157384)};
    \addplot [draw=none, fill=fp] coordinates {
        (0.25, 0.190474707928-0.2390412130194379)
        (0.35, 0.056849757547-0.0692370176138630)
        (0.45, 0.021427626304-0.0240480533546143)
        (0.55, 0.009923800776-0.0103745083509091)
        (0.65, 0.004164104639-0.0052300062703072)
        (0.75, 0.003448034309-0.0025508749928746)
        (0.85, 0.001875792931-0.0013466909878584)
        (0.95, 0.001042972003-0.0012825628455794)};
  \end{axis}
  \draw[draw=black] (0, 2.414) -- (3.32, 2.414);
\end{tikzpicture}
    \vspace{-4ex}
    \caption{}
    \label{pred_distribution}
  \end{subfigure}
  \hfill
  \begin{subfigure}{0.49\linewidth}
    \centering
    \input{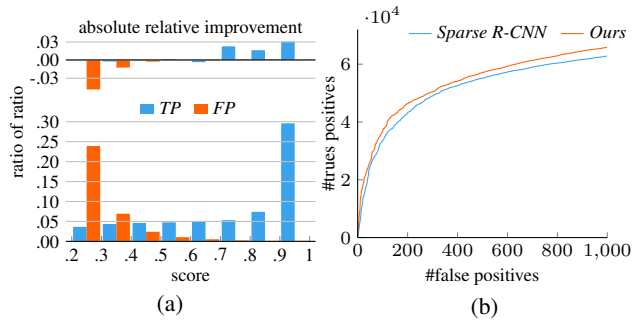}
    \vspace{-4ex}
    \caption{}
    \label{tp-fp_curve}
  \end{subfigure}
\vspace{-2ex}
\caption{~\ref{pred_distribution}. The bottom histogram describes the prediction distribution of Sparse RCNN~\cite{sun2020sparse} under different confidence scores, while the top one reflects the absolute improvements achieved by our approach compared with Sparse RCNN~\cite{sun2020sparse}. ~\ref{tp-fp_curve}. The $\text{FP-TP}$ curve when computing Average Precision (AP).}
\vspace{-1pc}
\label{fig:histogram}
\end{figure}

Recently, Carion et al. ~\cite{carion2020end} proposed a  fully end-to-end object detection framework DETR, which introduces learnable queries to represent objects and achieves competitive performance without any post-processing. It can be categorized as a \emph{query-based} approach to differentiate from the \emph{box-based}~\cite{lin2020focal, lin2017feature,2020atss} and \emph{point-based}~\cite{tian2019fcos, wang2020end} methods. Following DETR~\cite{carion2020end}, Sparse RCNN~\cite{sun2020sparse} ensures object queries interact with local feature of Region of Interest (RoI), while deformable DETR~\cite{zhu2021deformable} proposes attention modules that only attend to a small set of key sampling points. They further improve the detection accuracy and mitigate several issues occurred in DETR: slow convergence and high computational overhead.

The above success inspires us to study \emph{query-based} object detection methods in crowded scenes, aiming at designing a more sophisticated end-to-end detection framework. Although these ~\emph{query-based} approaches~\cite{gossipnet, zhu2021deformable} can obtain significant results on the slightly crowded datasets like COCO ~\cite{lin2014microsoft}, our initial studies show they suffer from several unresolved challenges in crowded scenes:(1). the \emph{query-based} detector tends to infer multiple predictions for a single object, with \emph{false positives} introduced. Figure.~\ref{fig:visulization}\textcolor{red}{a} shows a common failure case; (2). The performance of a \emph{query-based} detector becomes saturated or even worse as the depth of decoding stage increases, which is depicted in the Appendix.

\vspace{-0.5cm}
\paragraph{Our motivations.}

Further investigations on the \emph{query-based} method, Sparse RCNN~\cite{sun2020sparse}, yield the following intriguing findings in crowd scenes. 
As described in Figure.~\ref{pred_distribution}, a large percentage of target objects can be accurately predicted by those predictions with high confidence scores (e.g.\ higher than a threshold of 0.7), while containing few false positives. These predictions are more likely to be \emph{true positives} that can be taken as \emph{accepted predictions}. While the rest, where a considerable number of \emph{true positives} and \emph{false positives} exist, can be regarded as \emph{noisy predictions}.  Naturally, if an object is detected by one accepted prediction, there is no need for noisy predictions to detect it again. Hence, \emph{Why not strengthen the discrimination of those noisy predictions given the context of the accepted predictions}? To this end, the noisy queries can `perceive' whether their targets have been detected or not. If so, their confidence scores will be reduced and then filtered out.

\vspace{-0.5cm}
\paragraph{Our contributions}

Motivated by this, we propose a progressive prediction method equipped with a prediction selector, relation information extractor, query updater, and label assignment to improve the performance of query-based object detectors in handling crowded scenes. 

First, we develop a prediction selector to select queries associated with high confidence scores as \emph{accepted queries}, leaving the rest as \emph{noisy queries}. Then, to let the \emph{noisy queries} `perceive' whether their targets have been detected or not, we design a relation extractor for relation modeling between \emph{noisy queries} and their accepted neighbors. Further, a query updater is developed by performing a new local self-attention attending to spatially-related neighbors only. Finally, a new one-to-one label assignment rule is introduced to assign samples among the accepted and refined noisy queries step by step. With the proposed method, the above problems can be well addressed: (1). Each object can be detected only once, which greatly decreases the number of false positives while increasing the number of true positives, as described in Figure.~\ref{fig:visulization}\textcolor{red}{b}; (2). As depicted in Figure.~\ref{tp-fp_curve}, the performance is consistently improved compared with its counterparts~\cite{sun2020sparse,zhu2021deformable} that have the same depth of decoding stage.

Our method is generic and can be incorporated into multiple architectures~\cite{sun2020sparse, zhu2021deformable}, and delivers significant performance improvements of query-based detectors. Equipped with our approach, Sparse RCNN~\cite{sun2020sparse} with \emph{ResNet-50}~\cite{he2016deep} backbone obtains \textbf{92.0\%} $\text{AP}$, \textbf{41.4\%} $\text{MR}^{-2}$ and \textbf{83.2\%} $\text{JI}$ on the challenging dataset \emph{CrowdHuman}~\cite{shao2018crowdhuman}, outperforming the box-based method MIP~\cite{chu2020detection}. Besides, deformable DETR~\cite{zhu2021deformable}, equipped with our approach, also achieves \textbf{92.1\%} $\text{AP}$ and \textbf{84.0\%} $\text{JI}$. Moreover, our approach works reasonably well for less crowded scenes, e.g.\ the Sparse RCNN with our approach can still obtain \textbf{1.0\%} $\text{MR}^{-2}$ and \textbf{1.1\%} AP gains on moderately and slightly crowded datasets \emph{Citypersons}~\cite{zhang2017citypersons} and \emph{COCO}~\cite{lin2014microsoft}, respectively.

\vspace{-0.2cm}
\section{Related works}

\vspace{-0.2cm}
\paragraph{End-to-end object detection.}

RelationNet~\cite{hu2018relation} is one of the pioneering works trying to predict results directly, achieving promising performance compared to their counterparts on several famous benchmarks. DETR~\cite{carion2020end}  introduces learnable queries to represent objects and perform single prediction for each instance directly. Subsequently, deformable-DETR~\cite{zhu2021deformable}  limits the attention field of each query to a local area around the reference points to accelerate the convergence and improve detection performance. Meanwhile, Sparse R-CNN~\cite{sun2020sparse}  utilizes a fixed set of learnable queries to formulate objects instead of a number of proxy representation, e.g.\ anchors. Analogous to deformable DETR, RoIAlign~\cite{he2017mask} is applied to limit the attention field in a local region. Adaptive Clustering Transformer~\cite{act2021} proposes to improve the attention distribution in DETR’s encoder by LSH approximate clustering for convergence acceleration. UP-DETR~\cite{up2021detr} designs a new self-supervised method to improve the convergence speed of DETR, while TSP\cite{sun2020tsp} analyzes the main factors contributing to slow convergence in DETR. SMCA~\cite{fastCdetr} explores a better information interaction mechanism to further accelerate convergence and improve the performance of DETR.

\vspace{-0.5cm}
\paragraph{Object detection in crowded scenes.}

Research community has poured much interest in exploring occlusion problems on pedestrian detection. Specific methods have been proposed to mitigate this problem, including detecting by parts \cite{lu2019semantic, chi2020pedhunter, chi2020relational, zhang2019double,zhang2018occlusion} and improving hand-crafted rules in training target design. Recently, CNN-based methods have dominated the crowded object detection and achieved considerable gains. Several works propose new loss functions to address problems of crowded detection~\cite{wang2017repulsion,zhang2018occlusion}. Besides, the effectiveness of NMS is based on the assumption that multiple instances rarely occur at the same location in an image, which is not true in crowded scenes. But designing duplicate removal for crowded scenes is non-trivial. Soft-NMS \cite{bodla2017soft} and Softer-NMS \cite{he2018softer} replace hard removal of nearby proposals with score decay. Several works propose to use a neural network to simulate the function of NMS for duplicates removal~\cite{gossipnet, qi2018sequential}. Others explore NMS-aware training, including NMS with adaptive threshold \cite{hosang2016a,liu2019adaptive}, feature embedding \cite{salscheider2021featurenms} and multiple prediction with set suppression~\cite{chu2020detection, huang2020nms}, to tackle problem of object detection in crowded scenes.

Recently, PEDR~\cite{lin2020detr} proposes several techniques to improve the performance of \emph{query-based} detectors in coping with crowded detection, which is orthogonal to ours. Their techniques are also applicable to our work.

\vspace{-0.5cm}
\paragraph{Relation modeling for object detection.}
As discussed in ~\cite{hu2018relation}, early works~\cite{divvala, co-occurrent, torralba, auto-context, in_the_wild} use object relations as a post-processing step. The detected objects are re-scored by considering object relationships. For example, co-occurrence, which indicates how likely two object classes can exist in the same image, is used by DPM~\cite{dpm2010} to refine object scores. The subsequent approaches~\cite{a_role_of_context, tree_based} try more complex relation models, by taking additional positions and size into account. These methods achieve moderate success in the pre-deep learning era but do not prove the effectiveness in CNNs. Several recent works perform spatial reasoning ~\cite{acfobjdetection,end2endlstm,sptial_memory, gossipnet} to model object relations. Among them, GossipNet~\cite{gossipnet} and RelationNet~\cite{hu2018relation} are the representative methods. Both share the same spirit of modeling relations among boxes. However, the network of GossipNet~\cite{gossipnet} is complex (depth\textgreater 80) and its computation cost is demanding. Although it allows end-to-end learning in principle, no experimental evidence approves. RelationNet~\cite{hu2018relation} utilized the self-attention for feature interaction and obtained a promising improvement in general object detection. Nevertheless, it doesn't show a promising performance in dealing with crowd scenes~\cite{shao2018crowdhuman}.

Recent works related to ours are PS-RCNN~\cite{psrcnn} and IterDet~\cite{iterdet2021}. They proposed to detect objects according to the previous predicted ones. They need to mask the feature~\cite{psrcnn} or produce a history map~\cite{iterdet2021} to memorize the previous detections, introducing noise while limiting performance improvement~\cite{psrcnn} or incur heavy computation~\cite{iterdet2021}. Even so, both of them need a post-processing method to remove duplicates in every iteration. 

Recent query-based object detectors~\cite{sun2020sparse, zhu2021deformable, act2021, up2021detr, sun2020tsp, fastCdetr} utilized learnable queries to represent objects, and take advantage of the self/cross-attention to model the relations among queries, detecting objects in an end-to-end manner.  Our work inherits the methodology and boosts their performance in heavily, moderately, and slightly crowded scenes.

\section{Methodology}

In this section, we first revisit the query-based object detector, e.g.\ Sparse RCNN~\cite{sun2020sparse} briefly. Next, we illustrate our approach primarily deployed on Sparse RCNN explicitly. Finally, the main differences of detector design will be discussed as follows.

\subsection{Query Based Object Detector}
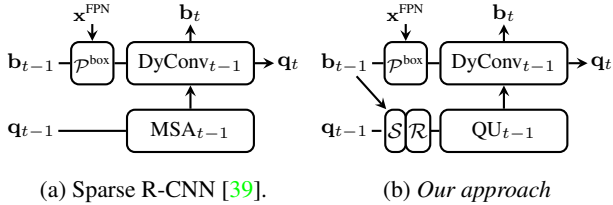
\begin{figure}[t]
\centering
\begin{subfigure}{0.49\linewidth}
\centering
\definecolor{brown}{HTML}{843C0C}
\definecolor{darkred}{HTML}{C43C0C}
\definecolor{skyblue}{HTML}{00B0F0}
\begin{tikzpicture}[
    scale=0.9,
    sym/.style={inner sep=1},
    box/.style={rectangle, rounded corners=3, draw=black, thick, minimum size=16, minimum width=48, fill=white},
    roi/.style={rectangle, rounded corners=3, draw=black, thick, minimum size=16, inner sep=0, fill=white},
    rnet/.style={rectangle, minimum size=16, minimum width=18},
    dr/.style={rectangle, rounded corners=3, draw=black, thick, minimum height=16, inner sep=1, fill=white},
    arr/.style={thick, ->, >=stealth, draw=black}
  ]
  \footnotesize
  \setlength{\medmuskip}{1\medmuskip}
  \setlength{\thickmuskip}{1\thickmuskip}

  \node[sym]   at (0, 0)       (prev_q)              {$\mathbf{q}_{t-1}$};
  \node[sym]   at (0, 1)       (prev_box)            {$\mathbf{b}_{t-1}$};
  \node[sym]   at (2.35, 1.75) (box)                 {$\mathbf{b}_{t}$};
  \node[sym]   at (3.8, 1)     (q)                   {$\mathbf{q}_{t}$};
  \node[box]   at (2.35, 1)     (dyn_conv)            {DyConv$_{t-1}$};
  
  \node[sym]   at (0.9, 1.75)      (x)                   {\scriptsize $\mathbf{x}^{\text{FPN}}$};
  \node[roi]   at (0.9, 1)      (roi)                 {\scriptsize $\mathcal{P}^{\text{box}}$};
  \draw[arr] (x)                -- (roi);

  \node[box]   at (2.35, 0)     (self_attn)           {MSA$_{t-1}$};

  \draw[arr] (prev_box) -- (roi) -- (dyn_conv) -- (box);
  \draw[arr] (prev_q)   -- (self_attn) -- (dyn_conv);
  \draw[arr] (dyn_conv) -- (q);

\end{tikzpicture}
\vspace{-2ex}
\caption{Sparse R-CNN~\cite{sun2020sparse}.}
\label{fig:sparse_rcnn_arch}
\end{subfigure}
\begin{subfigure}{0.49\linewidth}
\centering
\definecolor{brown}{HTML}{843C0C}
\definecolor{darkred}{HTML}{C43C0C}
\definecolor{skyblue}{HTML}{00B0F0}
\begin{tikzpicture}[
    scale=0.9,
    sym/.style={inner sep=1},
    box/.style={rectangle, rounded corners=3, draw=black, thick, minimum size=16, minimum width=48, fill=white},
    roi/.style={rectangle, rounded corners=3, draw=black, thick, minimum size=16, inner sep=0, fill=white},
    rnet/.style={rectangle, minimum size=16, minimum width=18},
    dr/.style={rectangle, rounded corners=3, draw=black, thick, minimum height=16, inner sep=1, fill=white},
    arr/.style={thick, ->, >=stealth, draw=black}
  ]
  \footnotesize
  \setlength{\medmuskip}{1\medmuskip}
  \setlength{\thickmuskip}{1\thickmuskip}

  \node[sym]   at (0, 0)       (prev_q)              {$\mathbf{q}_{t-1}$};
  \node[sym]   at (0, 1)       (prev_box)            {$\mathbf{b}_{t-1}$};
  \node[sym]   at (2.35, 1.75) (box)                 {$\mathbf{b}_{t}$};
  \node[sym]   at (3.8, 1)     (q)                   {$\mathbf{q}_{t}$};
  \node[box]   at (2.35, 1)     (dyn_conv)            {DyConv$_{t-1}$};
  
  \node[sym]   at (0.9, 1.75)      (x)                   {\scriptsize $\mathbf{x}^{\text{FPN}}$};
  \node[roi]   at (0.9, 1)      (roi)                 {\scriptsize $\mathcal{P}^{\text{box}}$};
  \draw[arr] (x)                -- (roi);

  \node[box]   at (2.35, 0)     (self_attn)           {QU$_{t-1}$};

  \node[dr, anchor=east] at (.9+.015, 0)  (d)             {$\mathcal{S}$};
  \node[dr, anchor=west] at (.9-.015, 0)  (r)             {$\mathcal{R}$};
  \node[rnet]            at (.9, 0)  (rnet)              {};

  \draw[arr] (prev_box) -- (roi) -- (dyn_conv) -- (box);
  \draw[arr] (prev_q)   -- (rnet) -- (self_attn) -- (dyn_conv);
  \draw[arr] (dyn_conv) -- (q);
  \draw[arr] (prev_box) -- (rnet);

\end{tikzpicture}
\vspace{-2ex}
\caption{\emph{Our approach}}
\label{fig:sr_sparse_rcnn}
\end{subfigure}
\vspace{-2ex}
\caption{The diagram of decoding stage. $\mathcal{P}$--RoIAlignPool~\cite{he2017mask}, $\text{DynConv}$ -- Dynamic Convolution, $\text{MSA}$ -- Multi-head Self-Attention, $\mathcal{S}$ -- Prediction Selector, $\mathcal{R}$ -- Relation Information Extractor, $\text{QU}$ -- Query Updater.}
\vspace{-1pc}
\end{figure}

Our approach can be deployed on most \emph{query-based} object detectors~\cite{carion2020end, sun2020sparse, zhu2021deformable}. To illustrate the proposed method, we choose Sparse RCNN~\cite{sun2020sparse} as our default instantiation. Figure.~\ref{fig:sparse_rcnn_arch} depicts its object detection pipeline, which can also be formulated as:

\begin{equation}
\begin{split}
    x_{t-1} &\gets \mathcal{P}^{box}(x^{FPN}, b_{t-1}), \\
    q^{\star}_{t-1} &\gets {\rm MSA}_{t-1}(q_{t-1}), \\
    q_{t} &\gets { \rm DynConv}_{t-1}(q^{\star}_{t-1}, x_{t-1}), \\
    b_t &\gets \mathcal{B}_{t-1}(q_t),
\end{split}
\label{equ:iterative_decoder}
\end{equation}
where $q \in R^{N \times d}$ denotes the learnable object query. $N$ and $d$ denote the number and dimension of query $q$, respectively. At stage $t$, an RoIAlign~\cite{he2017mask} $ \mathcal{P}^{box}$ extracts RoI features from FPN features $x^{FPN}$, under the guidance of bounding box $b_{t-1}$ predicted by the previous stage. Meanwhile, a multi-head self-attention module ${\rm MSA}_{t-1}$ is applied to the input query $q_{t-1}$ to get the transformed query $q^{\star}_{t-1}$. Then, a dynamic convolution module ${\rm DynConv}_{t-1}$ takes both $x_{t-1}$ and $q^{\star}_{t-1}$ as inputs and performs dynamic convolution to generates $q_{t}$ for the next stage. Simultaneously, $q_{t}$ is fed into the box prediction branch $\mathcal{B}_{t-1}$ for current bounding box prediction $b_t$, which is the input of the next stage $t$.

\begin{figure*}[htbp]
  \centering
  \includegraphics[width=1\textwidth]{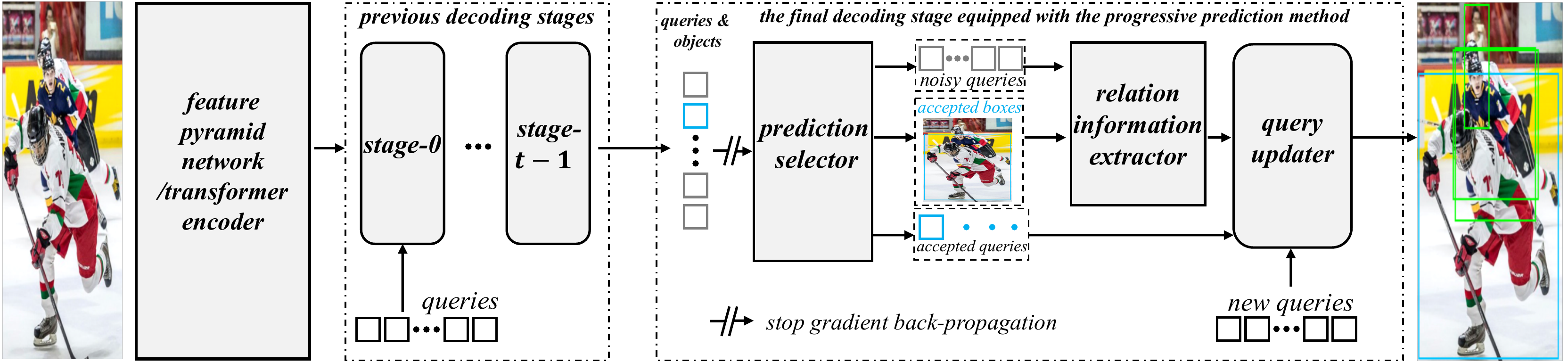}
\caption{The diagram of the proposed progressive end-to-end object detection framework. First, the \emph{prediction selector} select queries associated with high confidence scores as \emph{accepted queries}, leaving the rest as \emph{noisy queries}. Then \emph{Relation information extractor} models the relations between \emph{noisy queries} and their neighbors from accepted predictions. Next, the queries are fed into the \emph{queries updater} to be further refined by performing a new local self-attention.}
\label{fig:framwork}
\vspace{-1.5pc}
\end{figure*}

\subsection{Our Method}
\label{sec:approach_declearation}

As illustrated in Figure.~\ref{fig:framwork}, the proposed progressive predicting method consists of several components: prediction selector, relation information extractor, query updater, and label assignment, which will be introduced in detail next.

\vspace{-0.4cm}
\paragraph{Prediction selector.}

For the findings described in Sec.~\ref{intro},  a prediction selector is developed to select those queries prone to generating predictions with high confidence scores as accepted queries, while leaving the rest as noisy ones that need to be further refined. This procedure can be formulated in Equ.\eqref{equ:detection_split}.
\begin{equation}
\begin{split}
    \mathcal{D}^{h}_{t-1} &\gets \{{b_{i}| s_{i} \geq s \wedge b_{i} \in \mathcal{D}_{t-1}} \}, \\
    \mathcal{D}^{l}_{t-1} &\gets \mathcal{D}_{t-1} - \mathcal{D}^{h}_{t-1}. \\
\end{split}\label{equ:detection_split}    
\end{equation}
where $t$ is the stage number. $\mathcal{D}_{t-1}$ denote the whole predictions produced by the whole queries in the previous $t-1$ stage. $\mathcal{D}^h_{t-1}$ and $\mathcal{D}^l_{t-1}$ indicate the accepted predictions and noisy predictions generated from the accepted and noisy queries, respectively. $b_{i}$ and $s_{i}$ denote the predicted box and its confidence score, respectively. $s$ is the confidence score threshold.

\vspace{-0.4cm}
\paragraph{Relation information extractor.} 
As mentioned in Sec.~\ref{intro}, a  large percentage of target objects can be accurately predicted by the accepted queries. Therefore, if an object is detected by one accepted prediction, there is no need for noisy predictions to detect it again. In order to equip these noisy queries with the capability of perceiving whether their targets have been detected or not, we develop a relation information extractor to model the relation between the noisy predictions and their accepted neighbors.

\begin{figure}[t]
\centering
\definecolor{brown}{HTML}{843C0C}
\definecolor{darkred}{HTML}{C43C0C}
\definecolor{skyblue}{HTML}{00B0F0}

\begin{tikzpicture}[
    sym/.style={},
    box/.style={rectangle, rounded corners=3, draw=black, thick, minimum size=16},
    res/.style={circle, draw=black, thick, minimum size=12, inner sep=0, fill=white},
    arr/.style={thick, ->, >=stealth, draw=black}
  ]
  \small
  \node[sym, anchor=east]               at (0, 0)       (prev_q)      {$\mathbf{q}_{i}$};
  \node[sym, anchor=east, align=right]  at (0, 1.5)     (prev_box)    {$\mathbf{b}_{i},$\\$\mathcal{N}(\mathbf{b}_{i})$};
  \node[box]                            at (0.6, 1.5)   (eps)         {$\mathcal{E}$};
  \node[box]                            at (1.8, 1.5)   (r1)          {\textit{linear+r}};
  \node[box]                            at (3.25, 1.5)  (r2)          {\textit{linear}};
  \node[box]                            at (1.2, 0)     (r3)          {\textit{linear+r}};
  \node[box]                            at (2.85, 0)    (r4)          {\textit{linear}};
  \node[box]                            at (4, 0.75)    (pool)        {\textit{MaxPool}};
  \node[res]                            at (4, 0)       (add1)        {$+$};
  \node[box]                            at (5.1, 0)     (r5)          {$\mathcal{T}$};

  \draw[arr] (prev_box) -- (eps);
  \draw[arr] (eps)      -- (r1);
  \draw[arr] (r1)       -- (r2);
  \draw[arr] (r2)       -- (4, 1.5) -- (pool);
  \draw[arr] (pool)     -- (add1);

  \draw[arr] (prev_q) -- (r3);
  \draw[arr] (r3)     -- (r4);
  \draw[arr] (r4)     -- (add1);
  \draw[arr] (add1)   -- (r5);
  \draw[arr] (r5)   -- +(1, 0);

\end{tikzpicture}
\vspace{0.2cm}
\caption{Relation information extractor $\mathcal{R}$. $\mathcal{E}$ -- sine and cosine spatial positional encoding function~\cite{hu2018relation,attnyouneed}, \textit{linear} -- \textit{fc} layer, \textit{r} -- ReLU, $\mathcal{T}$-- \textit{fc} layer.}
\label{fig:relation_extractor}
\vspace{-0.08cm}
\end{figure}
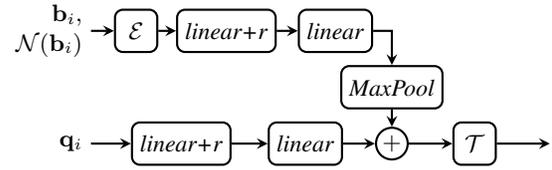

The detailed design of the relation information extractor is illustrated in Figure.~\ref{fig:relation_extractor}, with the procedure formulated in Equ.\eqref{equ:connection} as well. For each noisy prediction $b_i$, we first find their accepted neighbors $\mathcal{N}(b_i)$ in $\mathcal{D}^h_{t-1}$, constructing the spatially-related pairs $(b_i, \mathcal{N}(b_i))$. Then, the encoded pairs together with the \textit{intersection-over-union (IoU)} between them are fed to a compact network to obtain the geometry relation features $\mathcal{H}(b_i)$. Since the number of accepted neighbors corresponding to each noisy prediction is uncertain. An aggregation function is employed to reduce $\mathcal{H}(b_i)$ to the same feature dimension, while maintaining the permutation-invariance property. In our approach, we use \textit{max} pooling by default. Besides, the pooled geometry features, fused with the transformed query features, are further activated by a non-linear function.

\vspace{-5pt}
\begin{equation}
\begin{split}
    \mathcal{N}(b_i) &\gets \{b_j| \mathcal{O}(b_i, b_j) \geq \theta \},  b_i \in \mathcal{D}^l_{t-1}, b_j \in \mathcal{D}^h_{t-1}, \\
    \mathcal{H}(b_i) &\gets \mathcal{U}(\mathcal{E}(b_i, \mathcal{N}(b_i))), \quad b_i \in {D}^l_{t-1}, \\
    \mathcal{R}(b_i) &\gets \mathcal{T}(\textit{MaxPool}(\mathcal{H}(b_i)) + \mathcal{F}(q_i)).
\end{split}\label{equ:connection}    
\end{equation}
where $\mathcal{N} (\cdot)$ represents a function that finds neighbors for a box $b_i$ based on the \textit{IoU} $\mathcal{O}(\cdot, \cdot)$ with a threshold $\theta$. Here, we use it to find the accepted neighbors in $\mathcal{D}^{h}_{t-1}$ for the noisy predictions in $\mathcal{D}^l_{t-1}$. $\mathcal{E}(\cdot, \cdot)$ refers to the sine and cosine spatial positional encoding function which is the same as that~\cite{hu2018relation, attnyouneed}. Also, $\mathcal{U}(\cdot, \cdot)$ denotes a function used to generate geometry relation features $\mathcal{H}(b_i)$ from the encoded inputs. The noisy query $q_i$ corresponds to noisy prediction $b_i$ in $\mathcal{D}^l_{t-1}$, transformed by the function $\mathcal{F}(\cdot)$. The pooled geometry features and transformed query features $\mathcal{F}(q_i)$ are fused through element-wise summation, followed by a function $\mathcal{T}$ to produce the desired relation features $\mathcal{R}(b_i)$.

As depicted in Figure.~\ref{fig:relation_extractor}, $\mathcal{U}(\cdot, \cdot)$ consists of two consecutive ~\textit{fc} layers with \textit{ReLU}~\cite{maas2013rectifier} activation to increase the  non-linearity. Note that $\mathcal{F}(\cdot)$ and $\mathcal{U}(\cdot)$ share the same architecture, but are weight-independent. Here, the gradients of $q_i$ are stopped from back-propagating to the previous stages.

\vspace{-0.5cm}
\paragraph{Query updater.}
\label{para:query_updater}
To further refine the features of noisy queries, a query updater is developed, which is formulated in Equ.\eqref{equ:update_query}. Since the data distribution of $\mathcal{D}^{l}_{t-1}$ and $\mathcal{D}^{h}_{t-1}$ is different from that of $\mathcal{D}_{t-1}$, a new set of learnable queries is first introduced to complement the relation features through element-wise summation. Then the set of complemented noisy queries is taken as the input query $q_{t-1}$ to perform a new local self-attention ${\rm LMSA}_{t-1}$ and the subsequent dynamic convolution given in Equ.\eqref{equ:iterative_decoder}.

\begin{equation}
\begin{split}
        q_{t-1} &\gets \{\widetilde{q_i}|\widetilde{q_i}=\mathcal{R}(b_i) + e_i\}, \ b_i \in \mathcal{D}^l_{t-1}, e_i \in E  \\
        q^{\star}_{t-1} &\gets \text{LMSA}_{t-1}(q_{t-1}).
\end{split}\label{equ:update_query}
\end{equation}

Since object detection mainly focuses on the local region in an image. We design a new local self-attention module  ${\rm LMSA}_t$ to update the noisy query $q_{t-1}$.  It ensures each query only interacts with local neighbors instead of the whole queries over the full image. The local self-attention first finds those neighbors of each query based on the boxes' IoUs whose values are greater than 0. Then it performs the `qkv' mechanism in the same way as MSA. To this end, we perform self-attention locally instead of globally.

Different from the neighbor finding rule in ~\cite{lin2020detr}, we adopt the function $\mathcal{N}(\cdot)$ to select neighbors from $\mathcal{D}_{t-1}$ that are spatially related to $q_{t-1}$ in terms of \textit{IoU}. Note that, the new local self-attention ${\rm LMSA}_{t-1}$ is used to replace the ${\rm MSA}_{t-1}$ in Equ.\eqref{equ:iterative_decoder} for feature interaction.

\vspace{-0.5cm}
\paragraph{Label assignment}

Since accepted queries tend to generate true positive predictions, while the noisy ones involve a considerable number of true positives and false positives. Towards end-to-end object detection, we introduce a new one-to-one label assignment rule to assign samples step by step. Specifically, we first match the accepted predictions $\mathcal{D}^{h}_{t-1}$ with the ground truth set of objects. Then remove those targets that have been matched, and mainly consider the bipartite matching between noisy predictions $\mathcal{D}^{l}_{t-1}$ and the remaining ground truth set of objects. This matching process is described in Algorithm~\ref{algorithm:first}\footnote{The HungarianMatch operation in Algorithm~\ref{algorithm:first} is a combinatorial optimization approach that solves the assignment problem, which is commonly used in \cite{carion2020end, sun2020sparse, zhu2021deformable, wang2020end} for one-to-one label assignment.}, where the matching cost computation is slightly different from the original version \cite{sun2020sparse}. A spatial prior is adopted to compute the matching cost $\mathcal{C}$, that is, the center of bounding box $b_t$ needs to fall in the corresponding target box. Except for it, the formulation of the matching cost function is identical to the original work.

\begin{algorithm}[t] 
\caption{Label Assignment for $\mathcal{D}^l_{t}$.} 
\label{alg:yn} 
\begin{algorithmic}[1] 
\Require 
$\mathcal{D}^{l}_{t}$, $ \mathcal{D}^{h}_{t}$, $\mathcal{G}$; \\
 $\mathcal{D}^{l}_{t}$: results of $\mathcal{D}^{l}_{t-1}$ in Equ.\eqref{equ:detection_split} from stage $t$; \\
 $\mathcal{D}^{h}_{t}$:  results of $\mathcal{D}^{h}_{t-1}$ in Equ.\eqref{equ:detection_split} from stage $t$;\\
$\mathcal{G}$: target boxes.
\Ensure 
The matched predictions $\mathcal{M}^{l}_{D}$ and corresponding targets $\mathcal{M}^{l}_G$ after assignment.
\State Compute matching costs $\mathcal{C}^{h}_{t}$ between $\mathcal{D}^{h}_{t}$ and $\mathcal{G}$;
\State $\mathcal{M}^{h}_G, \mathcal{M}^{h}_{D} = \text{HungarianMatch}(\mathcal{D}^{h}_{t}, \mathcal{G}, \mathcal{C}^{h}_{t})$;

\State $\mathcal{G}^{l}_{t} = \mathcal{G} - \mathcal{M}^{h}_G$;
\State Compute matching costs $\mathcal{C}^{l}_{t}$ between $\mathcal{D}^{l}_{t}$ and $\mathcal{G}^{l}_{t}$;
\State $\mathcal{M}^{l}_G, \mathcal{M}^{l}_{D} = \text{HungarianMatch}(\mathcal{D}^{l}_{t}, \mathcal{G}^{l}_{t}, \mathcal{C}^{l}_{t})$;

\\ 
\Return $\mathcal{M}^{l}_G, \mathcal{M}^{l}_{D}$; 
\end{algorithmic}\label{algorithm:first}
\end{algorithm}

\vspace{-1pt}
\subsection{Difference of Detector Design}

Generally, our approach can be deployed on most query-based object detectors ~\cite{carion2020end, sun2020sparse, zhu2021deformable}. To illustrate the proposed method, we choose Sparse RCNN~\cite{sun2020sparse} as our default instantiation. It consists $t$ ($t=6$ by default) decoding stages, each of which performs prediction according to Equ.~\eqref{equ:iterative_decoder}. As described in Figure.~\ref{fig:framwork}, we keep the first $t-1$ decoding stages unchanged and only equip the last stage with the proposed method. Therefore, main differences lie in the last decoding stage, which will be described in the following.

\vspace{-0.6cm}
\paragraph{Architecture of the last stage.} 

As depicted in Figure.~\ref{fig:sr_sparse_rcnn}, the last decoding stage $t$ first employs a prediction selector $\mathcal{S}$ to split queries into accepted queries and noisy queries according to the confidence scores of their associated predictions. Then they are input to the relation information extractor $\mathcal{R}$ to extract the relation feature between the noisy predictions and their accepted neighbors. Finally, queries are fed into the query updater $\text{QU}$ to be further refined for recognition and localization.

\vspace{-0.6cm}
\paragraph{Box prediction}
Like~\cite{sun2020sparse}, a box regression branch is used for box prediction in the first $t-1$ stages. Differently, for the box prediction in the last stage, we directly use the identity mapping results from the previous $t-1$ stage both in the training and testing phase. This is because, at the latter layers, the predicted bounding boxes are less likely to fluctuate, which is observed in ~\cite{lin2020detr}. Meanwhile, the recognition branch remains the same as that of previous stages. 



\vspace{-0.6cm} 
\paragraph{Training loss}

We adopt the set prediction
loss adopted in ~\cite{sun2020sparse,zhu2021deformable} for training. For the samples to train stage $t$, we remove those samples from accepted predictions $\mathcal{D}^h_{t-1}$. to mitigate the class-imbalance issue, we follow the negative sample filtering mechanism~\cite{zhang2018refinedet} to early reject those well-classified negative queries whose confidence scores are lower than 0.05.

\vspace{-0.3cm}
\section{Experiments}
\label{sec:exp}
In this section, we evaluate our approach on heavily, moderately, and slightly crowded datasets~\cite{shao2018crowdhuman,zhang2017citypersons,lin2014microsoft} to demonstrate the generality of the proposed method in diverse scenarios.
\vspace{-0.6cm} 
\paragraph{Datasets.}

We adopt three datasets -- \emph{CrowdHuman} \cite{shao2018crowdhuman}, \emph{CityPersons} \cite{zhang2017citypersons} and \emph{COCO} \cite{lin2014microsoft} -- for comprehensive evaluations on heavily, moderately and slightly crowded situations, respectively. Table ~\ref{tbl:datasets} lists the ``instance density'' of each dataset.  Since our proposed approach mainly aims to improve crowded detections, we perform most of the comparisons and ablation experiments on \emph{CrowdHuman}. The evaluation experiments on \emph{Citypersons}~\cite{zhang2017citypersons} and \emph{COCO}~\cite{lin2014microsoft} are also conducted to suggest the proposed method is robust to crowdedness.

\begin{table}[t]
  \centering
  \begin{tabular}{l|c|c}
      \toprule
      Dataset & \# objects/img & \# overlaps/img  \\
      \hline
      CrowdHuman \cite{shao2018crowdhuman} & 22.64  & 2.40 \\
      CityPersons \cite{zhang2017citypersons} & 6.47  & 0.32 \\
      COCO$^*$ \cite{lin2014microsoft} & 9.34  & 0.015 \\
      \bottomrule
  \end{tabular}
  \vspace{-2pt}
  \caption{\emph{Instance density} of each dataset.  The threshold for overlap statistics is $\mathrm{IoU} > 0.5$.  *Averaged by the number of classes.}
  \label{tbl:datasets}
\vspace{-1.pc}
\end{table} 

\vspace{-1pc} 
\paragraph{Evaluation metrics.}

Following ~\cite{chu2020detection}, we mainly take three criteria: $\text{AP}$, $\text{MR}^{-2}$ and $\text{JI}$ as evaluation metrics. Generally, a larger $\text{AP}$, larger $\text{JI}$ and smaller $\text{MR}^{-2}$ indicates a better performance.

\vspace{-1pc} 
\paragraph{Implementation details.} 
Unless otherwise specified, we take Sparse RCNN~\cite{sun2020sparse} as our default instantiation, using standard \textit{ResNet-50}~\cite{he2016deep} pre-trained on ImageNet as backbone. We train our model with the Adam optimizer with a momentum of 0.9 and weight decay of 0.0001. Models are trained for 50, 000 iterations. The initial learning rate is 0.00005 and reduced by a factor of 0.1 at iteration 37,500. The last stage joints the optimization after 5,000 iterations of training. $\lambda_{cls}=2, \lambda_{L1}=5, \lambda_{giou}=2$. The default number of proposal boxes, proposal features, and stages are set to 500, 500, and 6, respectively. Additionally, The dimension of intermediate features in relationship extractor $\mathcal{R}$ is 256. The gradients are detached at proposal boxes from the second stage to stabilize training. Besides, those negative samples, whose \textit{intersection-over-area} (IoA) between any ignore region is higher than a threshold of $0.7$,  are not involved in training. Further, the hyper-parameters $s$ and $\theta$ are $0.7$ and $0.4$ by default in different query-based detectors~\cite{sun2020sparse,zhu2021deformable}. 

\subsection{Experiments on CrowdHuman}
CrowdHuman~\cite{shao2018crowdhuman} contains 15,000, 4,370 and 5,000 images for training, validation and test, respectively. For a fair comparison, we re-implement most of the involved models~\cite{lin2017feature,huang2020nms,lin2020focal,liu2019adaptive,2020atss,chu2020detection,tian2021fcos, wang2020end,carion2020end,zhu2021deformable, sun2020sparse,lin2020detr, gossipnet, hu2018relation}. Results are evaluated on the validation set, using the full-body annotations in the dataset.

\vspace{-1pc}
\paragraph{Main results.} We compare with mainstream object detectors, including \textit{box-based}: \textit{one-stage}~\cite{lin2020focal, 2020atss} , \textit{two-stages}~\cite{chu2020detection,lin2017feature,huang2020nms,liu2019adaptive}, and \textit{point-based}~\cite{tian2021fcos, wang2020end} as well as \textit{query-based}~\cite{carion2020end,zhu2021deformable, sun2020sparse,lin2020detr}. 

As shown in Table.~\ref{tbl:sota}, our approach outperforms these well-established detectors, achieving significant performance improvements over the \textit{box-based}, \textit{point-based}, and \textit{query-based} counterparts, illustrating the effectiveness of our approach in handling crowded scenes. Specifically, our method achieves 1.8\% $\text{AP}$ and 0.9\% $\text{JI}$ gains over the state-of-the-art \textit{box-based} approach MIP~\cite{chu2020detection}, which specializes in coping with crowded scenes.

The \textit{query-based} method Sparse RCNN~\cite{sun2020sparse}, equipped with the proposed method and 500 queries, can achieve 92.0\% $\text{AP}$, 41.4\% $\text{MR}^{-2}$ and 83.2\% $\text{JI}$ on the challenging CrowdHuman dataset~\cite{shao2018crowdhuman}, which is 1.3\%, 3.3\% and 1.8\% better than its counterpart -- original Sparse RCNN~\cite{sun2020sparse}. When increasing the number of queries to 750, our approach can still obtain a better performance of 92.5\% $\text{AP}$ and 83.3\% $\text{JI}$. This is because more queries can cover more patterns of objects in the image, such as scale, size, position, and other characteristics. Additionally, equipped with our approach, deformable DETR ~\cite{zhu2021deformable}\footnote{The detail implementation of deformable DETR with the proposed schema is illustrated in the Appendix.} can also obtain 2.2\% $\text{MR}^{-2}$ improvements over the original deformable DETR~\cite{zhu2021deformable}. Moreover, It also achieves 1.4\% $\text{AP}$ and 1.6\% $\text{JI}$ gains over the \textit{box-based} method MIP~\cite{chu2020detection}, demonstrating the effectiveness of our approach.
 
\begin{table}[ht]
	\centering
	\begin{tabular}{llccc}
		\toprule
		Method & \#Queries & AP & $\text{MR}^{-2}$  & JI \\
		\hline
		\hline
		\textit{box-based} \\
		RetinaNet~\cite{lin2020focal} & - & 85.3 & 55.1 & 73.7 \\
		ATSS~\cite{2020atss} & - & 87.0 & 51.1 & 75.9 \\
        ATSS~\cite{2020atss}+$\text{MIP}$~\cite{chu2020detection} & - & 88.7 & 51.6 & 77.0 \\
		FPN~\cite{lin2017feature}+NMS & - & 85.8 & 42.9 & 79.8 \\
		FPN~\cite{lin2017feature}+soft NMS & - & 88.2 & 42.9 & 79.8 \\
		FPN+MIP~\cite{chu2020detection} & - & 90.7 & 41.4 & 82.4 \\
		\hline
	    ${\text{FPN}}^{\dag}\text{+NMS}$ & - & 84.9 & 46.3 & -- \\
	    ${\text{Adaptive NMS}}^{\dag}$~\cite{liu2019adaptive} & - & 84.7 & 47.7 & -- \\
	    ${\text{PBN}^{\dag}}$~\cite{huang2020nms} & - & 89.3 & 43.4 & -- \\
		\hline
		\hline
		\textit{point-based} \\
		FCOS~\cite{tian2021fcos} & - & 86.8 & 54.0 & 75.7 \\
		FCOS~\cite{tian2021fcos}+$\text{MIP}$~\cite{chu2020detection} & - & 87.3 & 51.2 & 77.3 \\
		POTO~\cite{wang2020end} & - & 89.1 & 47.8 & 79.3 \\
		\hline
		\hline
		\textit{query-based} \\
		DETR~\cite{carion2020end} & 100 & 75.9 & 73.2 & 74.4 \\
		PEDR \cite{lin2020detr} & 1000 & 91.6 & 43.7 & 83.3 \\
		D-DETR \cite{zhu2021deformable} & 1000 & 91.5 & 43.7 & 83.1 \\
		S-RCNN~\cite{sun2020sparse} & 500 & 90.7 & 44.7 & 81.4 \\
		S-RCNN~\cite{sun2020sparse} & 750 & 91.3 & 44.8 & 81.3 \\
		\hline
		S-RCNN+\emph{Ours} & 500 & 92.0 & \textbf{41.4} & 83.2 \\
		S-RCNN+\emph{Ours} & 750 & \textbf{92.5} & 41.6 & 83.3 \\
		D-DETR+\emph{Ours} & 1000 & 92.1 & 41.5 & \textbf{84.0} \\
		\bottomrule
	\end{tabular}
	\caption{Comparisons of different methods on \emph{CrowdHuman} validation set, $\text{+MIP}$ represents multiple instance prediction with set NMS as post-processing. ${\dag}$ indicates the approach is implemented by PBM~\cite{huang2020nms}. S-RCNN -- Sparse RCNN~\cite{sun2020sparse}. D-DETR -- deformable DETR~\cite{zhu2021deformable}.}
	\label{tbl:sota}
	\vspace{-1pc}
\end{table}

\vspace{-0.5cm}
\paragraph{Ablation study of different modules.}
To explore the effectiveness of the proposed modules in Sec.~\ref{sec:approach_declearation}, we conduct extensive ablation study of the relation information extractor $\mathcal{R}$, local self-attention module $\text{LMSA}$ and the newly initialized embedding \emph{E}. All experiments are conducted on Sparse RCNN~\cite{sun2020sparse} with 500 queries, \emph{ResNet-50}~\cite{he2016deep} backbone and evaluated on \emph{CrowdHuman} dataset. Table.~\ref{tbl:module_analysis} shows that the relation information extractor $\mathcal{R}$ can obtain an improvement of 0.8\% $\text{AP}$, 1.7\% $\text{MR}^{-2}$ and 1.6\% $\text{JI}$. It indicates its effectiveness in reducing false positives and recalling false negatives. Moreover, when equipped with the new local self-attention $\text{LMSA}$, the performance on three evaluation metrics is further boosted, since the local self-attention can reduce duplicates effectively. Further, the newly initialized embeddings, aiming to approximate the new data distribution of noise predictions, can slightly improve $\text{MR}^{-2}$.

\vspace{-0.5cm}
\paragraph{Ablation study of hyper-parameter $\textbf{s}$.}
To analyze the effect of the confidence score threshold $s$, we first formulate the relation between detection boxes and target boxes in an image as a bipartite graph $\mathcal{G}=(\mathcal{V},\mathcal{E})$. It consists of a set $\mathcal{V} = \mathcal{D} \bigcup \mathcal{G}$ and nodes $\mathcal{E}$. $\mathcal{D}$ represents a set of predicted boxes whose scores are higher than the pre-defined score threshold, while $\mathcal{G}$ denotes the target boxes. An edge in $\mathcal{E}$ is defined as overlapping when the \emph{IoU} value, between a box in $\mathcal{D}$ and the other one from $\mathcal{G}$, is higher than 0.5 by default\footnote{Here, we follow the procedure to compute evaluation metric $\text{JI}$.}. Hence, the matching results can be acquired after applying the Hungarian Algorithm. As shown in Figure.~\ref{fig:histogram}, as the confidence score increases, the number of \textit{true positives} shows a clean upward trend while the number of ~\textit{false positives} decreases rapidly. Also, Figure.~\ref{fig:bar_score} depicts the performance our method can achieve under different values of $s$, where the performance increases slightly as $s$ increases. Thus, if not specific, we set $s$ to 0.7 by default.

\vspace{-0.5cm}
\paragraph{Ablation study of hyper-parameter \textbf{${\theta}$}.}
Here, we analyze the effect of the hyper-parameter \textit{intersection-over-union} threshold $\theta$. As discussed in Sec.~\ref{intro}, making sure a box candidate can `perceive' its neighbors helps a noisy query decide to decrease its confidence score or not, which is also the prerequisite for our method to work effectively. Different settings of \textit{intersection-over-union} (\emph{IoU}) threshold $\theta$ may affect the performance of the whole detector. We perform experiments on the \emph{CrowdHuman} dataset~\cite{shao2018crowdhuman} with $s$ frozen as $0.7$ while changing the value of $\theta$ linearly. From Figure.~\ref{fig:bar_theta}, we found our approach is robust to the change of \emph{IoU} threshold. This success may attribute to the good approximating feature of the newly designed components.

\vspace{-0.5cm}
\paragraph{Comparison with previous relation modeling works.}
To differentiate the previous works and ours, we evaluate several representative \textit{relation modeling} methods: RelationNet~\cite{hu2018relation}, GossipNet~\cite{gossipnet}, IterDet~\cite{iterdet2021}. RelationNet~\cite{hu2018relation} utilized self-attention modules to model the relations among different predictions. Meanwhile, GossipNet~\cite{gossipnet} uses several hand-designed relation blocks to explore the relationships among the predicted boxes, while IterDet~\cite{iterdet2021} iteratively infers predictions based on a historical map produced from the previous iteration. We re-implement its re-scoring version for RelationNet~\cite{hu2018relation}. For GossipNet\footnote{GossipNet:\textit{https://github.com/hosang/gossipnet}} and IterDet\footnote{IterDet:\textit{https://github.com/saic-vul/iterdet}}, we use their open-source implementations for evaluation. All models use FPN~\cite{lin2017feature} with \emph{ResNet-50}~\cite{he2016deep} as backbone, following the same training setting in ~\cite{gossipnet, hu2018relation,lin2017feature}. 

As shown in Table.~\ref{tbl:relation_modeling}, our approach shows better performance when compared with previous relation modeling works. Surprisingly, both RelationNet~\cite{hu2018relation} and GossipNet~\cite{gossipnet} suffer from a significant drop in $\text{AP}$ and $\text{MR}^{-2}$. It could attribute to the sub-optimal label assignment rule. Since both of them choose the prediction with the highest confidence score around one target as the correct box and take the rest as negatives. The predicted coordinates are not involved in computing loss, which might lead to the performance degradation in crowded scenes.

\begin{table*}[!t]
\vspace{0mm}
\centering
\subfloat[Ablations of different modules.\label{tbl:module_analysis}]{
\begin{minipage}{.3\textwidth}
\centering
    \tablestyle{1pc}{1.05}\setlength{\tabcolsep}{1.mm}
    \begin{tabular}{ccc|ccc}
		\toprule
		  $\mathcal{R}$ & $\text{LMSA}$ & \emph{E} & AP & $\text{MR}^{-2}$  & JI \\
		\hline
		 & & & 90.7 & 44.7 & 81.4 \\
		\hline 
		\checkmark & & & 91.5 & 43.0 & 83.0  \\
		\checkmark & \checkmark &  & \textbf{92.0} & 42.0 & \textbf{83.5}  \\
		\checkmark & \checkmark & \checkmark & \textbf{92.0} & \textbf{41.4} &  83.2  \\
		\bottomrule
	\end{tabular}
	\vspace{0.4cm}
\end{minipage}
}
\subfloat[Comparisons of different relation modeling approaches.\label{tbl:relation_modeling}]{
\begin{minipage}{.32\textwidth}
\centering
\tablestyle{5pt}{1.05}\setlength{\tabcolsep}{1.mm}
	\begin{tabular}{l|c|ccc}
		\toprule
		Method & \#Queries & AP & $\text{MR}^{-2}$  & JI \\
		\hline
		GossipNet~\cite{gossipnet} & - & 80.4 & 49.4 & 81.6 \\
		RelationNet~\cite{hu2018relation} & - & 81.6 & 48.2 & 74.6 \\
		IterDet~\cite{iterdet2021} & - & 88.0 & 47.5 & 78.0 \\
		\hline
		D-DETR+\emph{Ours} & 500 & 91.2 & 42.6 & \textbf{84.0} \\
		S-RCNN+\emph{Ours} & 500 & \textbf{92.0} & \textbf{41.4} & 83.2 \\
		\bottomrule
	\end{tabular}
	\vspace{0.25cm}
\end{minipage}
}\hspace{0.3cm}
\subfloat[Performance comparisons on \emph{CityPersons}.~\label{tbl:citypersons_eval}]{
\begin{minipage}{0.32\textwidth}
\centering
\tablestyle{5pt}{1.05}\setlength{\tabcolsep}{1.mm}
    \begin{tabular}{l|c|cccc}
  \toprule
  Method & \#Queries & $\text{MR}^{-2}$ & AP \\
  \hline
  FPN+NMS & & 9.8 & 94.7 \\
  FPN+Soft-NMS~\cite{bodla2017soft} & - & 9.9 & 94.9  \\
  \text{MIP}~\cite{chu2020detection} &  & 8.8 & 95.8   \\
  D-DETR~\cite{zhu2021deformable} & 500 & 9.4 & 96.6 \\
  $\text{S-RCNN}$~\cite{sun2020sparse} & 500 & 10.0 & 96.8 \\
  \hline
  D-DETR+\emph{Ours} & 500 & 7.8 & 96.7 \\
  S-RCNN+\emph{Ours} & 500 & \textbf{7.8} & \textbf{97.6} \\
  \bottomrule
  \end{tabular}
  \end{minipage}
  }\vspace{-6pt}
  \caption{~\ref{tbl:module_analysis}. Ablation study of different modules proposed in our approach, taking Sparse RCNN~\cite{sun2020sparse} with 500 queries as our default instantiation. ~\ref{tbl:relation_modeling}. Comparisons of different relation modeling appraoches. All the experiments are conducted on ~\emph{CrowdHuman}~\cite{shao2018crowdhuman} dataset. ~\ref{tbl:citypersons_eval} Performance comparisons of different methods on ~\emph{CityPersons}~\cite{zhang2017citypersons}. Both \textit{box-based} ~\cite{chu2020detection,lin2017feature} and \textit{query-based} approaches~\cite{sun2020sparse,zhu2021deformable} are evaluated.}
  \vspace{-1pc}
\end{table*}

\begin{figure*}
\centering
\scriptsize
\begin{minipage}{0.35\textwidth}
  \centering
  \begin{tikzpicture}[
    box/.style={rectangle, rounded corners=2, draw=gray, fill=gray!5, align=center, minimum height=26},
  ]
  \centering
  \begin{axis}[
      ybar, axis on top,
      height=5cm, width=1.12\linewidth,
      bar width=.225cm,
      ybar=0,
      ymajorgrids, tick align=inside,
      major grid style={draw=gray!50},
      enlarge x limits=.15,
      ymin=0, ymax=3e4,
      axis x line*=bottom,
      y axis line style={opacity=0},
      major tick length=2pt,
      ytick style={draw=none},
      legend style={
          draw=none,
          fill=none,
          at={(1, 1.08)},
          inner sep=1.2,
          anchor=north east,
          legend columns=-1,
          /tikz/every even column/.append style={column sep=0.1cm}
        },
      legend image code/.code={\draw [#1, draw=none] (0cm,-0.05cm) rectangle (0.18cm,0.08cm); },
      ylabel=Number of errors (\#errors),
      ylabel near ticks,
      symbolic x coords={Duplicate,Localization,Background,Missing},
      xtick=data,
      ytick={0, 5e3, 1e4, 1.5e4, 2e4}
    ]
    \addplot [draw=none, fill=ji] coordinates {
      (Duplicate, 17798)
      (Localization, 14666)
      (Background, 3515)
      (Missing, 9948) };
    \addplot [draw=none,fill=ap] coordinates {
      (Duplicate, 9022)
      (Localization, 11396)
      (Background, 2607)
      (Missing, 9948) };

    \legend{Sparse R-CNN, ours}
    \node[box] at (axis cs:Duplicate, 2.48e4)    (_5) {\color{ji}{1.780}\\\color{ap}{0.902}};
    \node[box] at (axis cs:Localization, 2.48e4) (_6) {\color{ji}{1.467}\\\color{ap}{1.140}};
    \node[box] at (axis cs:Background, 2.48e4)   (_7) {\color{ji}{0.352}\\\color{ap}{0.261}};
    \node[box] at (axis cs:Missing, 2.48e4)      (_8) {\color{ji}{0.995}\\\color{ap}{0.995}};
  \end{axis}
\end{tikzpicture}
  \vspace{-6ex}
  \caption{Error analysis on Sparse RCNN~\cite{sun2020sparse} and our approach, with \emph{ResNet-50}~\cite{he2016deep} as backbone. The bar plots show different error types that contribute to the \textit{false positives}.}
  \label{fig:tide_decompositon}
\end{minipage}
\hfill
\begin{minipage}{0.62\textwidth}
\begin{subfigure}{0.43\textwidth}
  \centering
  \begin{tikzpicture}[
    box/.style={rectangle, rounded corners=2, draw=gray, fill=gray!5, align=center, minimum height=26},
  ]
  \centering
  \begin{axis}[
      ybar, axis on top,
      height=5cm, width=1.12\linewidth,
      bar width=0.15cm,
      ybar=0,
      ymajorgrids,
      tick align=inside,
      enlarge x limits=.1,
      axis x line*=bottom,
      y axis line style={opacity=0},
      ymin=0, ymax=1.5,
      major tick length=2pt,
      ytick style={draw=none},
      legend style={
          draw=none,
          fill=none,
          at={(1, 1.08)},
          inner sep=1.2,
          anchor=north east,
          legend columns=-1,
          /tikz/every even column/.append style={column sep=0.1cm}
        },
      legend image code/.code={\draw [#1, draw=none] (0cm,-0.05cm) rectangle (0.18cm,0.08cm); },
      ytick={0.0, 0.2, 0.4, 0.6, 0.8, 1.0},
      yticklabels={0.0, 0.2, 0.4, 0.6, 0.8, 1.0},
      xtick={0.5,0.6,0.7,0.8,0.9},
      xticklabels={0.5,0.6,0.7,0.8,0.9},
      ylabel near ticks,
      xlabel near ticks,
    ]
    \addplot [draw=none, fill=mr] coordinates {
        (0.5, 0.4242)
        (0.6, 0.4274)
        (0.7, 0.4140)
        (0.8, 0.4097)
        (0.9, 0.4150) };
    \addplot [draw=none, fill=ji] coordinates {
        (0.5, 0.8312)
        (0.6, 0.8326)
        (0.7, 0.8320)
        (0.8, 0.8330)
        (0.9, 0.8346) };
    \addplot [draw=none, fill=ap] coordinates {
        (0.5, 0.9195)
        (0.6, 0.9200)
        (0.7, 0.9210)
        (0.8, 0.9199)
        (0.9, 0.9186) };
    \legend{MR$^{-2}$, JI, AP}
    \node[box] at (axis cs:0.5, 1.24) (_5) {\color{mr}{.424}\\\color{ji}{.831}\\\color{ap}{.920}};
    \node[box] at (axis cs:0.6, 1.24) (_6) {\color{mr}{.427}\\\color{ji}{.833}\\\color{ap}{.920}};
    \node[box] at (axis cs:0.7, 1.24) (_7) {\color{mr}{.414}\\\color{ji}{.832}\\\color{ap}{.921}};
    \node[box] at (axis cs:0.8, 1.24) (_8) {\color{mr}{.410}\\\color{ji}{.834}\\\color{ap}{.920}};
    \node[box] at (axis cs:0.9, 1.24) (_9) {\color{mr}{.415}\\\color{ji}{.834}\\\color{ap}{.919}};
  \end{axis}
\end{tikzpicture}
  \vspace{-1ex}
  \caption{\textit{confidence score threshold} $s$}
  \label{fig:bar_score}
\end{subfigure}
\hfill
\begin{subfigure}{0.56\textwidth}
  \centering
  \begin{tikzpicture}[
    box/.style={rectangle, rounded corners=2, draw=gray, fill=gray!5, align=center, minimum height=26},
  ]
  \centering
  \begin{axis}[
      ybar, axis on top,
      height=5cm, width=1.12\linewidth,
      bar width=0.15cm,
      ybar=0,
      ymajorgrids,
      tick align=inside,
      enlarge x limits=.1,
      axis x line*=bottom,
      y axis line style={opacity=0},
      ymin=0, ymax=1.5,
      major tick length=2pt,
      ytick style={draw=none},
      legend style={
          draw=none,
          fill=none,
          at={(1, 1.08)},
          inner sep=1.2,
          anchor=north east,
          legend columns=-1,
          /tikz/every even column/.append style={column sep=0.1cm}
        },
      legend image code/.code={\draw [#1, draw=none] (0cm,-0.05cm) rectangle (0.18cm,0.08cm); },
      ytick={0.0, 0.2, 0.4, 0.6, 0.8, 1.0},
      yticklabels={0.0, 0.2, 0.4, 0.6, 0.8, 1.0},
      xtick={0.1,0.2,0.3,0.4,0.5,0.6,0.7},
      xticklabels={0.1,0.2,0.3,0.4,0.5,0.6,0.7},
      ylabel near ticks,
      xlabel near ticks,
    ]
    \addplot [draw=none, fill=mr] coordinates {
        (0.1, 0.4249)
        (0.2, 0.4272)
        (0.3, 0.4306)
        (0.4, 0.414)
        (0.5, 0.4222)
        (0.6, 0.4179)
        (0.7, 0.4231) };
    \addplot [draw=none, fill=ji] coordinates {
        (0.1, 0.8346)
        (0.2, 0.8324)
        (0.3, 0.8336)
        (0.4, 0.832)
        (0.5, 0.8357)
        (0.6, 0.8351)
        (0.7, 0.8354) };
    \addplot [draw=none, fill=ap] coordinates {
        (0.1, 0.9191)
        (0.2, 0.9177)
        (0.3, 0.9188)
        (0.4, 0.921)
        (0.5, 0.9201)
        (0.6, 0.9200)
        (0.7, 0.9199) };
    \legend{MR$^{-2}$, JI, AP}
    \node[box] at (axis cs:.1, 1.24) (_1) {\color{mr}{.425}\\\color{ji}{.835}\\\color{ap}{.919}};
    \node[box] at (axis cs:.2, 1.24) (_2) {\color{mr}{.427}\\\color{ji}{.833}\\\color{ap}{.918}};
    \node[box] at (axis cs:.3, 1.24) (_3) {\color{mr}{.431}\\\color{ji}{.834}\\\color{ap}{.919}};
    \node[box] at (axis cs:.4, 1.24) (_4) {\color{mr}{.414}\\\color{ji}{.832}\\\color{ap}{.921}};
    \node[box] at (axis cs:.5, 1.24) (_5) {\color{mr}{.422}\\\color{ji}{.836}\\\color{ap}{.920}};
    \node[box] at (axis cs:.6, 1.24) (_6) {\color{mr}{.418}\\\color{ji}{.835}\\\color{ap}{.920}};
    \node[box] at (axis cs:.7, 1.24) (_7) {\color{mr}{.423}\\\color{ji}{.835}\\\color{ap}{.920}};
  \end{axis}
\end{tikzpicture}
  \vspace{-1ex}
  \caption{\textit{intersection over union threshold} $\theta$}
  \label{fig:bar_theta}
\end{subfigure}
\caption{Performance of the proposed method with different configurations of hyper-parameter $\textbf{s}$ and \textbf{${\theta}$} on \emph{CrowdHuman}~\cite{shao2018crowdhuman} dataset.}
\end{minipage}
\vspace{-1pc}
\end{figure*}

\vspace{-0.5cm}
\paragraph{Analysis on false positives.}
To understand the factors contributing to the performance improvement, we conduct an error analysis on our method. We adopt the recently proposed TIDE~\cite{tide-eccv2020} to compare our approach with the counterpart Sparse RCNN~\cite{sun2020sparse}. We analyzed the composite error at Recall=0.9 for all methods. As illustrated in Figure.~\ref{fig:tide_decompositon}, our method performs better at removing duplication, providing more accurate localization, and reducing mistaken recognition. Since part of queries can perceive whether their targets are detected or not through the relation information extractor. Also, the local self-attention module ensures queries only interact with their neighbors rather than the whole. To this end, the duplicates could be eliminated efficiently. Besides, with identity mapping plugged in the last regression branch for box prediction, the number of training samples in the previous decoding stage increases, making the optimization much easier. Additionally, benefiting from the new learnable embeddings for data distribution approximation, the representation ability of object queries are further enhanced.

\vspace{-0.2cm}
\subsection{Experiments on Citypersons}

CityPersons~\cite{zhang2017citypersons} is one of the widely used benchmarks for pedestrian detection. It contains 5, 000 images (2, 975 for training, 500 for validation, and 1, 525 for testing, respectively). Each image has a size of 1024 $\times$ 2048. To improve the overall performance, we proposed to pre-train all models on the \emph{CrowdHuman} dataset and fine-tune them on ~\emph{CityPersons} (\emph{reasonable}) training subset, then tested on the (\emph{reasonable}) validation subset. For those ~\emph{box-based} methods, we train and evaluate them with the image resolution enlarged by 1.3$\times$ compared to the original one for better accuracy. The \emph{query-based} approaches are trained and evaluated at the original image size with 500 queries. The other settings remain the same as those of Sparse RCNN~\cite{sun2020sparse} and deformable DETR~\cite{zhu2021deformable}.

\vspace{-0.2cm}
\subsection{Experiments on COCO.}
According to Table ~\ref{tbl:datasets}, the crowdness of \emph{COCO}~\cite{lin2014microsoft} is
very low, which is beyond our design purpose. Nevertheless, we still conduct an experiment on this dataset to verify: 1) whether our method generalizes well to multi-class detection; 2) whether our approach can still handle slightly crowded scenarios, especially with \emph{isolated} instances.

Following the common practice of Sparse RCNN ~\cite{sun2020sparse} with 300 queries, we use a subset of 5000 images in the original validation set (named \emph{minival}) for validation while using the remaining images in the training and validation set for training. Except for the proposed modules and label assignment rule in the last stage, other settings remain the same as the original methods~\cite{zhu2021deformable, sun2020sparse}. Table.~\ref{tbl:mscoco_eval} shows the performance comparisons with deformable DETR~\cite{zhu2021deformable} and Sparse RCNN~\cite{sun2020sparse}. Moderate improvements are obtained, e.g. \textbf{}{0.9\%} $\text{AP}$ higher than the deformable DETR~\cite{zhu2021deformable} and \textbf{1.1\%} $\text{AP}$ higher than the Sparse RCNN~\cite{sun2020sparse}. The experimental results reflect the effectiveness of our progressive predicting approach in slightly crowded scenarios, proving the proposed method can also solve the performance saturation problem of \emph{query-based} detectors.

\begin{table}[t]
  \centering
  \begin{tabular}{p{23mm}|p{5mm}<{\centering}p{5mm}<{\centering}p{6mm}<{\centering}|p{5mm}<{\centering}p{5mm}<{\centering}p{5mm}<{\centering}}
  \toprule
  Method & AP & $\text{AP}_{50}$ & $\text{AP}_{75}$ &  $\text{AP}_{S}$ & $\text{AP}_{M}$  & $\text{AP}_{L}$\\
  \hline
  S-RCNN~\cite{sun2020sparse} & 45.0 & 64.2 & 49.1 & 27.6 & 47.5 & 59.1 \\
  D-DETR~\cite{zhu2021deformable} & 45.8 & 64.5 & 49.4 & 28.2 & 49.0 & 61.7 \\
  \hline
  S-RCNN+\emph{Ours} & 46.1 & 65.3 & 50.6 & 29.2  &  48.7  & 59.9 \\
  D-DETR+\emph{Ours} &  \textbf{46.7} & 65.3 & 50.3 & 28.6  &  49.8  & 61.7 \\
  \bottomrule
  \end{tabular}
  \caption{Performance comparisons of different methods on \emph{COCO} 2017 ~\cite{lin2014microsoft} \emph{minival} set.}
  \label{tbl:mscoco_eval}
 \vspace{-1pt}
\end{table}

\vspace{-0.2cm}
\section{Conclusion}
In this paper, we propose a progressive prediction method to boost the performance of \emph{query-based} object detectors in handling crowded scenes. Equipped with our approach, two representatives \textit{query-based} methods, Sparse RCNN~\cite{sun2020sparse} and deformable DETR~\cite{zhu2021deformable} achieve consistent improvements over the heavily, moderately, as well as slightly crowded datasets~\cite{shao2018crowdhuman, zhang2017citypersons, lin2014microsoft}, which suggests our approach is robust to crowdedness. Since Sparse RCNN~\cite{sun2020sparse} and deformable DETR~\cite{zhu2021deformable} require large computing resources, making it difficult for our method to be deployed on devices with limited computing capacity. How to develop a computation-efficacy end-to-end detector is still under exploration. Besides, we found the decision boundary for the noisy queries is unclear. We believe that the performance can be further improved if a better feature engineering method or loss function is adopted. However, it is beyond the purpose of this work.

\clearpage

  {\small
    \bibliographystyle{ieee_fullname}
    \bibliography{ref21, ref22}

\begin{thebibliography}{10}\itemsep=-1pt

\bibitem{bodla2017soft}
Navaneeth {Bodla}, Bharat {Singh}, Rama {Chellappa}, and Larry~S. {Davis}.
\newblock Soft-nms -- improving object detection with one line of code.
\newblock {\em arXiv preprint arXiv:1704.04503}, 2017.

\bibitem{tide-eccv2020}
Daniel Bolya, Sean Foley, James Hays, and Judy Hoffman.
\newblock Tide: A general toolbox for identifying object detection errors.
\newblock In {\em ECCV}, 2020.

\bibitem{carion2020end}
Nicolas {Carion}, Francisco {Massa}, Gabriel {Synnaeve}, Nicolas {Usunier},
  Alexander {Kirillov}, and Sergey {Zagoruyko}.
\newblock End-to-end object detection with transformers.
\newblock In {\em European Conference on Computer Vision}, pages 213--229,
  2020.

\bibitem{sptial_memory}
Xinlei Chen and Abhinav Gupta.
\newblock Spatial memory for context reasoning in object detection.
\newblock {\em CoRR}, abs/1704.04224, 2017.

\bibitem{chi2020pedhunter}
Cheng {Chi}, Shifeng {Zhang}, Junliang {Xing}, Zhen {Lei}, Stan {Li}, and
  Xudong {Zou}.
\newblock Pedhunter: Occlusion robust pedestrian detector in crowded scenes.
\newblock In {\em AAAI 2020 : The Thirty-Fourth AAAI Conference on Artificial
  Intelligence}, 2020.

\bibitem{chi2020relational}
Cheng {Chi}, Shifeng {Zhang}, Junliang {Xing}, Zhen {Lei}, Stan {Li}, and
  Xudong {Zou}.
\newblock Relational learning for joint head and human detection.
\newblock In {\em AAAI 2020 : The Thirty-Fourth AAAI Conference on Artificial
  Intelligence}, 2020.

\bibitem{tree_based}
Myung~Jin Choi, Antonio Torralba, and Alan~S. Willsky.
\newblock A tree-based context model for object recognition.
\newblock {\em IEEE Transactions on Pattern Analysis and Machine Intelligence},
  34(2):240--252, 2012.

\bibitem{chu2020detection}
Xuangeng Chu, Anlin Zheng, Xiangyu Zhang, and Jian Sun.
\newblock Detection in crowded scenes: One proposal, multiple predictions.
\newblock pages 12214--12223, 2020.

\bibitem{up2021detr}
Zhigang Dai, Bolun Cai, Yugeng Lin, and Junying Chen.
\newblock Up-detr: Unsupervised pre-training for object detection with
  transformers.
\newblock In {\em Proceedings of the IEEE/CVF Conference on Computer Vision and
  Pattern Recognition (CVPR)}, pages 1601--1610, June 2021.

\bibitem{divvala}
Santosh~K. Divvala, Derek Hoiem, James~H. Hays, Alexei~A. Efros, and Martial
  Hebert.
\newblock An empirical study of context in object detection.
\newblock In {\em 2009 IEEE Conference on Computer Vision and Pattern
  Recognition}, pages 1271--1278, 2009.

\bibitem{dpm2010}
Pedro~F. Felzenszwalb, Ross~B. Girshick, David McAllester, and Deva Ramanan.
\newblock Object detection with discriminatively trained part-based models.
\newblock {\em IEEE Transactions on Pattern Analysis and Machine Intelligence},
  32(9):1627--1645, 2010.

\bibitem{co-occurrent}
Carolina Galleguillos, Andrew Rabinovich, and Serge Belongie.
\newblock Object categorization using co-occurrence, location and appearance.
\newblock In {\em 2008 IEEE Conference on Computer Vision and Pattern
  Recognition}, pages 1--8, 2008.

\bibitem{fastCdetr}
Peng Gao, Minghang Zheng, Xiaogang Wang, Jifeng Dai, and Hongsheng Li.
\newblock Fast convergence of {DETR} with spatially modulated co-attention.
\newblock {\em CoRR}, abs/2101.07448, 2021.

\bibitem{psrcnn}
Zheng Ge, Zequn Jie, Xin Huang, Rong Xu, and Osamu Yoshie.
\newblock {PS-RCNN:} detecting secondary human instances in a crowd via primary
  object suppression.
\newblock {\em CoRR}, abs/2003.07080, 2020.

\bibitem{he2017mask}
Kaiming {He}, Georgia {Gkioxari}, Piotr {Dollár}, and Ross {Girshick}.
\newblock Mask r-cnn.
\newblock {\em computer vision and pattern recognition}, 2017.

\bibitem{he2016deep}
Kaiming {He}, Xiangyu {Zhang}, Shaoqing {Ren}, and Jian {Sun}.
\newblock Deep residual learning for image recognition.
\newblock In {\em 2016 IEEE Conference on Computer Vision and Pattern
  Recognition (CVPR)}, pages 770--778, 2016.

\bibitem{he2018softer}
Yihui {He}, Xiangyu {Zhang}, Marios {Savvides}, and Kris {Kitani}.
\newblock Softer-nms: Rethinking bounding box regression for accurate object
  detection.
\newblock 2018.

\bibitem{gossipnet}
J. Hosang, R. Benenson, and B. Schiele.
\newblock Learning non-maximum suppression.
\newblock In {\em CVPR}, 2017.

\bibitem{hosang2016a}
Jan~Hendrik {Hosang}, Rodrigo {Benenson}, and Bernt {Schiele}.
\newblock A convnet for non-maximum suppression.
\newblock In {\em 38th German Conference on Pattern Recognition}, pages
  192--204, 2016.

\bibitem{hu2018relation}
Han {Hu}, Jiayuan {Gu}, Zheng {Zhang}, Jifeng {Dai}, and Yichen {Wei}.
\newblock Relation networks for object detection.
\newblock In {\em 2018 IEEE/CVF Conference on Computer Vision and Pattern
  Recognition}, pages 3588--3597, 2018.

\bibitem{huang2020nms}
Xin Huang, Zheng Ge, Zequn Jie, and Osamu Yoshie.
\newblock Nms by representative region: Towards crowded pedestrian detection by
  proposal pairing.
\newblock In {\em Proceedings of the IEEE/CVF Conference on Computer Vision and
  Pattern Recognition}, pages 10750--10759, 2020.

\bibitem{kingma2014adam}
Diederik~P Kingma and Jimmy Ba.
\newblock Adam: A method for stochastic optimization.
\newblock {\em arXiv preprint arXiv:1412.6980}, 2014.

\bibitem{acfobjdetection}
Jianan Li, Yunchao Wei, Xiaodan Liang, Jian Dong, Tingfa Xu, Jiashi Feng, and
  Shuicheng Yan.
\newblock Attentive contexts for object detection.
\newblock {\em IEEE Transactions on Multimedia}, 19(5):944--954, 2017.

\bibitem{lin2020detr}
Matthieu {Lin}, Chuming {Li}, Xingyuan {Bu}, Ming {Sun}, Chen {Lin}, Junjie
  {Yan}, Wanli {Ouyang}, and Zhidong {Deng}.
\newblock Detr for crowd pedestrian detection.
\newblock {\em arXiv preprint arXiv:2012.06785}, 2020.

\bibitem{lin2017feature}
Tsung-Yi {Lin}, Piotr {Dollar}, Ross {Girshick}, Kaiming {He}, Bharath
  {Hariharan}, and Serge {Belongie}.
\newblock Feature pyramid networks for object detection.
\newblock In {\em 2017 IEEE Conference on Computer Vision and Pattern
  Recognition (CVPR)}, pages 936--944, 2017.

\bibitem{lin2020focal}
Tsung-Yi {Lin}, Priya {Goyal}, Ross {Girshick}, Kaiming {He}, and Piotr
  {Dollar}.
\newblock Focal loss for dense object detection.
\newblock {\em IEEE Transactions on Pattern Analysis and Machine Intelligence},
  42(2):318--327, 2020.

\bibitem{lin2014microsoft}
Tsung-Yi {Lin}, Michael {Maire}, Serge~J. {Belongie}, James {Hays}, Pietro
  {Perona}, Deva {Ramanan}, Piotr {Dollár}, and C.~Lawrence {Zitnick}.
\newblock Microsoft coco: Common objects in context.
\newblock In {\em European Conference on Computer Vision}, pages 740--755,
  2014.

\bibitem{liu2019adaptive}
Songtao {Liu}, Di {Huang}, and Yunhong {Wang}.
\newblock Adaptive nms: Refining pedestrian detection in a crowd.
\newblock In {\em 2019 IEEE/CVF Conference on Computer Vision and Pattern
  Recognition (CVPR)}, pages 6459--6468, 2019.

\bibitem{liu2021swin}
Ze Liu, Yutong Lin, Yue Cao, Han Hu, Yixuan Wei, Zheng Zhang, Stephen Lin, and
  Baining Guo.
\newblock Swin transformer: Hierarchical vision transformer using shifted
  windows.
\newblock {\em arXiv preprint arXiv:2103.14030}, 2021.

\bibitem{lu2019semantic}
Ruiqi {Lu} and Huimin {Ma}.
\newblock Semantic head enhanced pedestrian detection in a crowd.
\newblock {\em arXiv preprint arXiv:1911.11985}, 2019.

\bibitem{maas2013rectifier}
Andrew~L Maas, Awni~Y Hannun, Andrew~Y Ng, et~al.
\newblock Rectifier nonlinearities improve neural network acoustic models.
\newblock In {\em Proc. icml}, volume~30, page~3. Citeseer, 2013.

\bibitem{in_the_wild}
Roozbeh Mottaghi, Xianjie Chen, Xiaobai Liu, Nam-Gyu Cho, Seong-Whan Lee, Sanja
  Fidler, Raquel Urtasun, and Alan Yuille.
\newblock The role of context for object detection and semantic segmentation in
  the wild.
\newblock In {\em 2014 IEEE Conference on Computer Vision and Pattern
  Recognition}, pages 891--898, 2014.

\bibitem{a_role_of_context}
Roozbeh Mottaghi, Xianjie Chen, Xiaobai Liu, Nam-Gyu Cho, Seong-Whan Lee, Sanja
  Fidler, Raquel Urtasun, and Alan Yuille.
\newblock The role of context for object detection and semantic segmentation in
  the wild.
\newblock In {\em 2014 IEEE Conference on Computer Vision and Pattern
  Recognition}, pages 891--898, 2014.

\bibitem{qi2018sequential}
Lu {Qi}, Shu {Liu}, Jianping {Shi}, and Jiaya {Jia}.
\newblock Sequential context encoding for duplicate removal.
\newblock In {\em Advances in Neural Information Processing Systems},
  volume~31, pages 2049--2058, 2018.

\bibitem{iterdet2021}
Danila Rukhovich, Konstantin Sofiiuk, Danil Galeev, Olga Barinova, and Anton
  Konushin.
\newblock Iterdet: Iterative scheme for objectdetection in crowded
  environments.
\newblock {\em CoRR}, abs/2005.05708, 2020.

\bibitem{salscheider2021featurenms}
Niels~Ole {Salscheider}.
\newblock Featurenms: Non-maximum suppression by learning feature embeddings.
\newblock In {\em 2020 25th International Conference on Pattern Recognition
  (ICPR)}, pages 7848--7854, 2021.

\bibitem{shao2018crowdhuman}
Shuai {Shao}, Zijian {Zhao}, Boxun {Li}, Tete {Xiao}, Gang {Yu}, Xiangyu
  {Zhang}, and Jian {Sun}.
\newblock Crowdhuman: A benchmark for detecting human in a crowd.
\newblock {\em arXiv preprint arXiv:1805.00123}, 2018.

\bibitem{end2endlstm}
Russell Stewart, Mykhaylo Andriluka, and Andrew~Y. Ng.
\newblock End-to-end people detection in crowded scenes.
\newblock In {\em 2016 IEEE Conference on Computer Vision and Pattern
  Recognition (CVPR)}, pages 2325--2333, 2016.

\bibitem{sun2020sparse}
Peize {Sun}, Rufeng {Zhang}, Yi {Jiang}, Tao {Kong}, Chenfeng {Xu}, Wei {Zhan},
  Masayoshi {Tomizuka}, Lei {Li}, Zehuan {Yuan}, Changhu {Wang}, and Ping
  {Luo}.
\newblock Sparse r-cnn: End-to-end object detection with learnable proposals.
\newblock {\em arXiv preprint arXiv:2011.12450}, 2020.

\bibitem{sun2020tsp}
Zhiqing Sun, Shengcao Cao, Yiming Yang, and Kris Kitani.
\newblock Rethinking transformer-based set prediction for object detection.
\newblock {\em CoRR}, abs/2011.10881, 2020.

\bibitem{tian2019fcos}
Zhi {Tian}, Chunhua {Shen}, Hao {Chen}, and Tong {He}.
\newblock Fcos: Fully convolutional one-stage object detection.
\newblock In {\em 2019 IEEE/CVF International Conference on Computer Vision
  (ICCV)}, pages 9626--9635, 2019.

\bibitem{tian2021fcos}
Zhi Tian, Chunhua Shen, Hao Chen, and Tong He.
\newblock {FCOS}: A simple and strong anchor-free object detector.
\newblock 2021.

\bibitem{torralba}
Torralba, Murphy, Freeman, and Rubin.
\newblock Context-based vision system for place and object recognition.
\newblock In {\em Proceedings Ninth IEEE International Conference on Computer
  Vision}, pages 273--280 vol.1, 2003.

\bibitem{auto-context}
Zhuowen Tu.
\newblock Auto-context and its application to high-level vision tasks.
\newblock In {\em 2008 IEEE Conference on Computer Vision and Pattern
  Recognition}, pages 1--8, 2008.

\bibitem{attnyouneed}
Ashish Vaswani, Noam Shazeer, Niki Parmar, Jakob Uszkoreit, Llion Jones,
  Aidan~N Gomez, \L~ukasz Kaiser, and Illia Polosukhin.
\newblock Attention is all you need.
\newblock In I. Guyon, U.~V. Luxburg, S. Bengio, H. Wallach, R. Fergus, S.
  Vishwanathan, and R. Garnett, editors, {\em Advances in Neural Information
  Processing Systems}, volume~30. Curran Associates, Inc., 2017.

\bibitem{wang2020end}
Jianfeng {Wang}, Lin {Song}, Zeming {Li}, Hongbin {Sun}, Jian {Sun}, and
  Nanning {Zheng}.
\newblock End-to-end object detection with fully convolutional network.
\newblock {\em arXiv preprint arXiv:2012.03544}, 2020.

\bibitem{wang2017repulsion}
Xinlong {Wang}, Tete {Xiao}, Yuning {Jiang}, Shuai {Shao}, Jian {Sun}, and
  Chunhua {Shen}.
\newblock Repulsion loss: Detecting pedestrians in a crowd.
\newblock {\em arXiv preprint arXiv:1711.07752}, 2017.

\bibitem{zhang2019double}
Kevin {Zhang}, Feng {Xiong}, Peize {Sun}, Li {Hu}, Boxun {Li}, and Gang {Yu}.
\newblock Double anchor r-cnn for human detection in a crowd.
\newblock {\em arXiv preprint arXiv:1909.09998}, 2019.

\bibitem{zhang2017citypersons}
Shanshan {Zhang}, Rodrigo {Benenson}, and Bernt {Schiele}.
\newblock Citypersons: A diverse dataset for pedestrian detection.
\newblock In {\em 2017 IEEE Conference on Computer Vision and Pattern
  Recognition (CVPR)}, pages 4457--4465, 2017.

\bibitem{2020atss}
Shifeng {Zhang}, Cheng {Chi}, Yongqiang {Yao}, Zhen {Lei}, and Stan~Z. {Li}.
\newblock Bridging the gap between anchor-based and anchor-free detection via
  adaptive training sample selection.
\newblock {\em arXiv preprint arXiv:1912.02424}, 2019.

\bibitem{zhang2018occlusion}
Shifeng {Zhang}, Longyin {Wen}, Xiao {Bian}, Zhen {Lei}, and Stan~Z. {Li}.
\newblock Occlusion-aware r-cnn: Detecting pedestrians in a crowd.
\newblock In {\em Proceedings of the European Conference on Computer Vision
  (ECCV)}, pages 637--653, 2018.

\bibitem{zhang2018refinedet}
Shifeng Zhang, Longyin Wen, Xiao Bian, Zhen Lei, and Stan~Z. Li.
\newblock Single-shot refinement neural network for object detection.
\newblock In {\em CVPR}, 2018.

\bibitem{act2021}
Minghang Zheng, Peng Gao, Xiaogang Wang, Hongsheng Li, and Hao Dong.
\newblock End-to-end object detection with adaptive clustering transformer.
\newblock {\em CoRR}, abs/2011.09315, 2020.

\bibitem{zhu2021deformable}
Xizhou {Zhu}, Weijie {Su}, Lewei {Lu}, Bin {Li}, Xiaogang {Wang}, and Jifeng
  {Dai}.
\newblock Deformable detr: Deformable transformers for end-to-end object
  detection.
\newblock In {\em ICLR 2021: The Ninth International Conference on Learning
  Representations}, 2021.

\end{thebibliography}
  }
\newpage
\appendix
\vspace{-0.4cm}
\section{Implementation of Deformable DETR with progressive predicting method.}

\begin{figure}[!t]
  \centering
  \begin{subfigure}{\linewidth}
    \definecolor{brown}{HTML}{843C0C}
\definecolor{darkred}{HTML}{C43C0C}
\definecolor{skyblue}{HTML}{00B0F0}
\definecolor{black}{HTML}{000000}

\begin{tikzpicture}[
    scale=0.9,
    sym/.style={inner sep=1},
    box/.style={rectangle, rounded corners=3, draw=black, thick, minimum size=16, minimum width=54},
    roi/.style={rectangle, rounded corners=3, draw=black, thick, minimum size=16, inner sep=0, fill=white},
    rnet/.style={rectangle, minimum size=16},
    dr/.style={rectangle, rounded corners=3, draw=black, thick, minimum height=16, inner sep=1, fill=white},
    arr/.style={thick, ->, >=stealth, draw=black}
  ]
  \small

  \node[sym]   at (0, 0)       (prev_q)              {$\mathbf{q}_{t-1}$};
  \node[sym]   at (0, 3)       (prev_box)            {$\mathbf{b}_{t-1}$};
  \node[box]   at (2, 1.5)     (dyn_conv)            {MCA$_{t-1}$};
  \node[box]   at (2, 0)       (self_attn)           {MSA$_{t-1}$};
  \node[sym]   at (2, 3)       (box)                 {$\mathbf{b}_{t}$};
  \node[sym]   at (.2+4, 0)    (q)                   {$\mathbf{q}_{t}$};
  \node[box]   at +(.4+6, 1.5) (_dyn_conv)           {MCA$_t$};
  \node[box]   at +(.4+6, 0)   (_self_attn)          {MSA$_t$};
  \node[sym]   at +(.4+8, 0)   (_q)                  {$\mathbf{q}_{t+1}$};
  \node[sym]   at +(.4+6, 3)   (_box)                {$\mathbf{b}_{t+1}$};

  \draw[arr] (prev_box)         -- (dyn_conv.west);
  \draw[arr] (prev_q)           -- (self_attn);
  \draw[arr] (self_attn)        -- (dyn_conv);
  \draw[arr] (dyn_conv)         -- (box);
  \draw[arr] (dyn_conv.east)    -- (q);
  \draw[arr] (box) -- (.2+4, 3) -- (_dyn_conv.west);
  \draw[arr] (_self_attn)       -- (_dyn_conv);
  \draw[arr] (_dyn_conv)        -- (_box);
  \draw[arr] (_dyn_conv.east)   -- (_q);
  \draw[arr] (q)                -- (_self_attn);
\end{tikzpicture}
    \caption{Decoder in deformable DETR~\cite{zhu2021deformable}. $\text{MCA}$ -- multi-head cross-attention, $\text{MSA}$ -- multi-head self-attention.}
    \label{fig:ddetr_arch}
  \end{subfigure}
  \vspace{-0.5pc}
  \begin{subfigure}{\linewidth}
    \definecolor{brown}{HTML}{843C0C}
\definecolor{darkred}{HTML}{C43C0C}
\definecolor{skyblue}{HTML}{00B0F0}
\definecolor{black}{HTML}{000000}
\begin{tikzpicture}[
    scale=0.9,
    sym/.style={inner sep=1},
    box/.style={rectangle, rounded corners=3, draw=black, thick, minimum size=16, minimum width=54},
    roi/.style={rectangle, rounded corners=3, draw=black, thick, minimum size=16, inner sep=0, fill=white},
    rnet/.style={rectangle, minimum size=16},
    dr/.style={rectangle, rounded corners=3, draw=black, thick, minimum height=16, inner sep=1, fill=white},
    arr/.style={thick, ->, >=stealth, draw=black}
  ]
  \small
  \node[sym]             at (0, 0)       (prev_q)        {$\mathbf{q}_{t-1}$};
  \node[sym]             at (0, 3)       (prev_box)      {$\mathbf{b}_{t-1}$};
  \node[box]             at (2, 1.5)     (dyn_conv)      {MCA$_{t-1}$};
  \node[box]             at (2, 0)       (self_attn)     {MSA$_{t-1}$};
  \node[sym]             at (2, 3)       (box)           {$\mathbf{b}_{t}$};
  \node[box, draw=none]  at (2, 3)       (hidden_box)    {};
  \node[sym]             at (.2+4, 0)    (q)             {$\mathbf{q}_{t}$};
  \node[box]             at +(.4+6, 1.5) (_dyn_conv)     {MCA$_t$};
  \node[box]             at +(.4+6, 0)   (_self_attn)    {LMSA$_t$};
  \node[sym]             at +(.4+8, 0)   (_q)            {$\mathbf{q}_{t+1}$};
  \node[sym]             at +(.4+6, 3)   (_box)          {$\mathbf{b}_{t+1}$};
  \node[dr, anchor=east] at (.2+4+.01, 1.5)  (d)             {$\mathcal{S}$};
  \node[dr, anchor=west] at (.2+4-.01, 1.5)  (r)             {$\mathcal{R}$};
  \node[rnet]            at (.2+4, 1.5)  (rnet)          {};

  \draw[arr] (prev_box)         -- (dyn_conv.west);
  \draw[arr] (prev_q)           -- (self_attn);
  \draw[arr] (self_attn)        -- (dyn_conv);
  \draw[arr] (dyn_conv)         -- (box);
  \draw[arr] (dyn_conv.east)    -- (q);
  \draw[arr] (box) -- (.2+4, 3) -- (_dyn_conv.west);
  \draw[arr] (hidden_box.east)  -- (rnet);
  \draw[arr] (rnet)             -- (_self_attn.west);
  \draw[arr] (_self_attn)       -- (_dyn_conv);
  \draw[arr] (_dyn_conv)        -- (_box);
  \draw[arr] (_dyn_conv.east)   -- (_q);
  \draw[arr] (q)                -- (.2+4, 1.2);
\end{tikzpicture}
    \caption{Decoder in SR-Deformable DETR (Ours). $\mathcal{S}$ -- Prediction Selector, $\mathcal{R}$ -- Relation information extractor, $\text{LMSA}$ -- local multi-head self-attention.}
    \label{fig:sr_ddetr}
  \end{subfigure}
  \vspace{-0.5pc}
  \caption{~\ref{fig:ddetr_arch} is the architecture of decoding stage in deformable DETR~\cite{zhu2021deformable};
  ~\ref{fig:sr_ddetr} describes the decoding stage structure equipped with our designed components for progressive predicting schema. }
  
\end{figure}
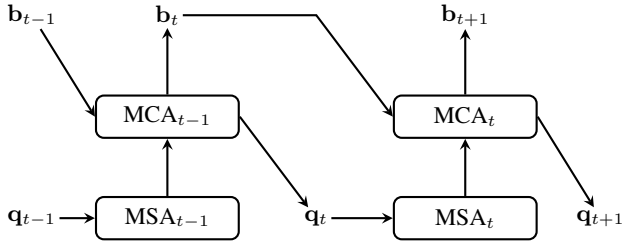
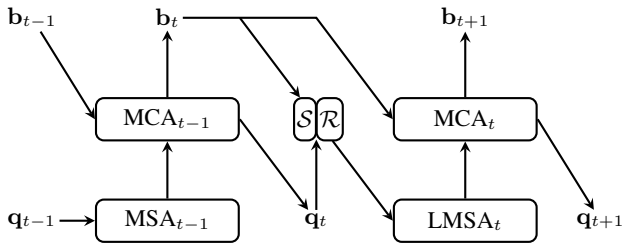

We also deploy our progressive predicting approach on the deformable DETR~\cite{zhu2021deformable} to demonstrate the generality of our method. Similar to Sparse RCNN ~\cite{sun2020sparse}, the decoder in deformable DETR consists of 6 stages, which is depicted in Figure.~\ref{fig:ddetr_arch}. As described in the paper, we integrate our designed components into the last decoding stage. Figure.~\ref{fig:sr_ddetr} also depicts its detail architecture. For the hyper-parameters setting, i.e\ confidence score threshold $s$, are identical to those adopted in Sparse RCNN~\cite{sun2020sparse}.

We choose the deformable DETR with iterative bounding box refinement. Following deformable DETR~\cite{zhu2021deformable}, we use \emph{ResNet-50}~\cite{he2016deep} as backbone. The whole detector is trained with an Adam optimizer~\cite{kingma2014adam} and a weight decay of 0.0001. The total training duration is 50 epochs on 8 GPUs with 1 image per GPU. The initial learning rate is 0.0002 and dropped by a factor of 0.1 after 40 epochs. The parameters initialization in the newly added components and losses weights are identical to the original work~\cite{zhu2021deformable}. The default number of queries and stages is 500 and 6, respectively. The hyper-parameters $s$ and $\theta$ are also $0.7$ and $0.4$, respectively. The gradients are detached at proposal boxes from the second stage to stabilize training. We stop gradient back-propagation from the last stage to the previous ones. Besides, those negative samples that overlap with ignore region with an \textit{intersection-over-area}(IoA) greater than $0.7$ are not involved in training.


\vspace{-0.2cm}

\vspace{-0.2cm}
\section{Performance change of a query-based decoder when handling crowded scenes.}
The performance of a \emph{query-based} detector would not be improved but will degrade as the depth of a decoder increases when handling crowded scenes. Experiments are conducted on \emph{CrowdHuman} dataset, taking Sparse RCNN based on \emph{ResNet-50} as base detector. It equips with 500 queries. We adjust the depth of its decoder while keeping the others unchanged. As is described in Table. ~\ref{tbl:depth_exp}, the performance degrades as the depth of the decoder increases.

\vspace{-0.2cm}
\begin{table}[ht]
	\centering
	\begin{tabular}{l|c|ccc}
		\toprule
		 \#Depth& \#Queries & AP & $\text{MR}^{-2}$  & JI \\
		\hline
		 6 & & 90.7 & 44.7 & 81.4 \\
		 7 & & 90.6 & 45.7 & 81.0 \\
		8 & 500& 90.4 & 45.9 & 80.3 \\
		9 & & 90.7 & 44.4 & 80.9 \\
        10 & & 90.2 & 46.6 & 80.0 \\
		\bottomrule
	\end{tabular}
	\caption{Experiment analysis as the depth of a decoder increases, which performs on \emph{CrowdHuman} dataset.}
	\label{tbl:depth_exp}
	\vspace{-1.5pc}
\end{table}

\vspace{-0.1cm}
\section{Performance of query detector with large model in crowded scenes.}

To explore the detection upper bound of a query-based detector in tackling crowded scenes, we replace the \emph{ResNet-50} with a large backbone, Swin-Large~\cite{liu2021swin}. Experiments are conducted on \emph{CrowdHuman}~\cite{shao2018crowdhuman} and \emph{CityPersons}~\cite{zhang2017citypersons} datasets, with the same training strategy described in the paper. As depicted in Table.~\ref{tbl:large_model}, our method can significantly boost the performance of a query-based detector, which achieves a \textit{state-of-the-arts} results on both \emph{CrowdHuman} and \emph{CityPersons}  validating datasets.

\vspace{-0.2cm}
\begin{table}[ht]
	\centering
	\begin{tabular}{p{22mm}|p{12mm}<{\centering}|p{11mm}<{\centering}|p{4mm}<{\centering}p{5mm}<{\centering}p{4mm}<{\centering}}
		\toprule
		 Method& Dataset & \#Queries & AP & $\text{MR}^{-2}$  & JI \\
		\hline
		 S-RCNN&  & 500 & 93.1 & 39.9 & 85.1 \\
		 S-RCNN+\emph{Ours} &\emph{CHuman} & 500 &  93.8 & \textbf{37.4} & 86.5 \\
		D-DETR & & 1000& 93.4 & 39.6 & 86.3 \\
		D-DETR+\emph{Ours} & & 1000& \textbf{94.1} & 37.7 & \textbf{87.1} \\
		\hline
		S-RCNN&  & 500 & 98.3 & 5.9 & 93.7 \\
		 D-DETR &\emph{CPersons} & 500 &  96.4 & 8.4 & 92.0 \\
		S-RCNN+\emph{Ours} & & 500& \textbf{98.4} & \textbf{4.9} & \textbf{94.2} \\
		D-DETR+\emph{Ours} & & 500& 97.5 & 5.9 & 93.7 \\
		
		\bottomrule
	\end{tabular}
	\caption{Experiment on \emph{CHuman}(\emph{CrowdHuman}) and \emph{CPersons}(\emph{CityPersons}) with Swin-L~\cite{liu2021swin}. S-RCNN -- Sparse RCNN~\cite{sun2020sparse}, D-DETR -- Deformable DETR~\cite{zhu2021deformable}}
	\label{tbl:large_model}
	\vspace{-2pc}
\end{table}
\clearpage

\newpage
\begin{figure*}
  \small
  \centering
  \setlength{\tabcolsep}{3pt}
\begin{tabular}{cccccc}
  \toprule
  RelationNet~\cite{hu2018relation} & IterDet~\cite{iterdet2021} & Sparse R-CNN~\cite{sun2020sparse} & D-DETR~\cite{carion2020end} & Sparse RCNN+Ours & D-DETR+Ours \\
  \midrule

  \includegraphics[width=0.15\textwidth,trim={0 0 0 2.4cm},clip]{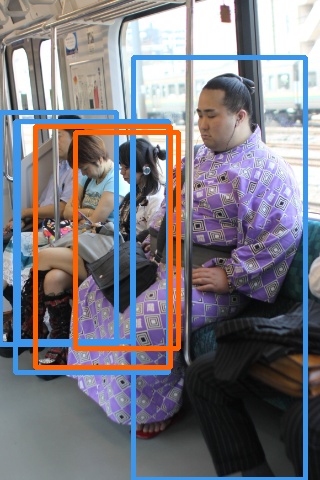} &
  \includegraphics[width=0.15\textwidth,trim={0 0 0 2.4cm},clip]{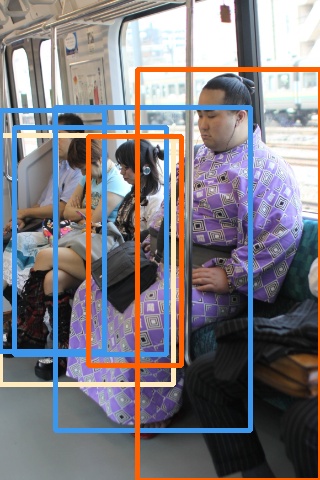} &
  \includegraphics[width=0.15\textwidth,trim={0 0 0 2.4cm},clip]{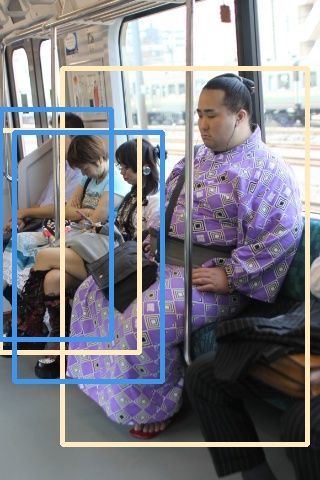} &
  \includegraphics[width=0.15\textwidth,trim={0 0 0 2.4cm},clip]{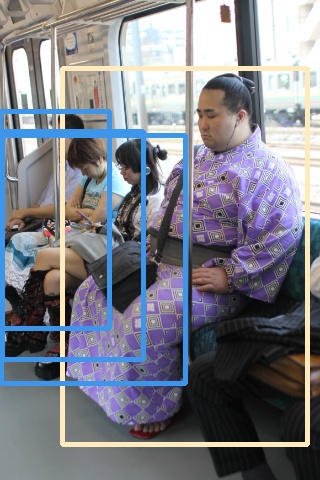} &
  \includegraphics[width=0.15\textwidth,trim={0 0 0 2.4cm},clip]{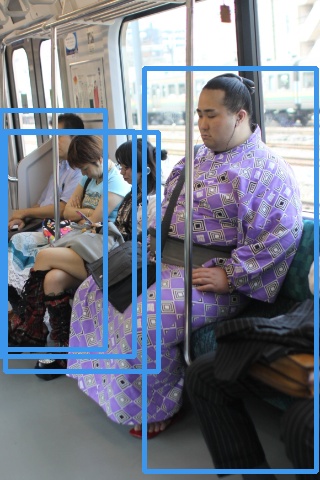} &
  \includegraphics[width=0.15\textwidth,trim={0 0 0 2.4cm},clip]{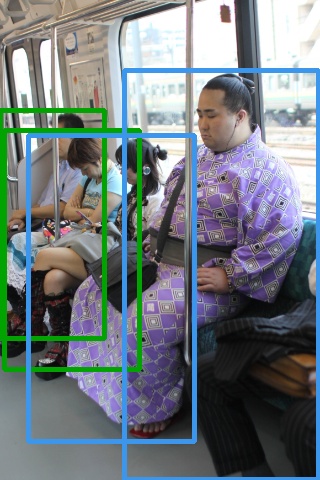} \\

  \includegraphics[width=0.15\textwidth]{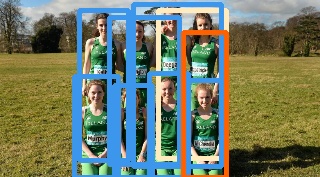} &
  \includegraphics[width=0.15\textwidth]{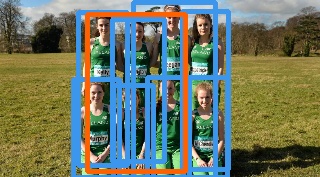} &
  \includegraphics[width=0.15\textwidth]{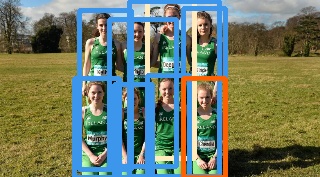} &
  \includegraphics[width=0.15\textwidth]{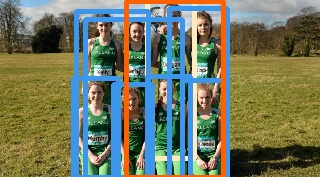} &
  \includegraphics[width=0.15\textwidth]{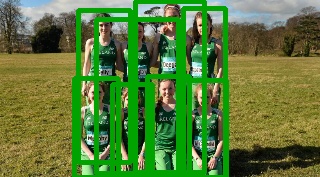} &
  \includegraphics[width=0.15\textwidth]{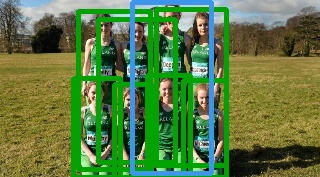} \\
  
  \includegraphics[width=0.15\textwidth,trim={0 1.1cm 0 0},clip]{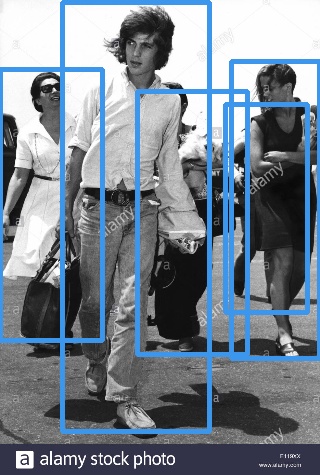} &
  \includegraphics[width=0.15\textwidth,trim={0 1.1cm 0 0},clip]{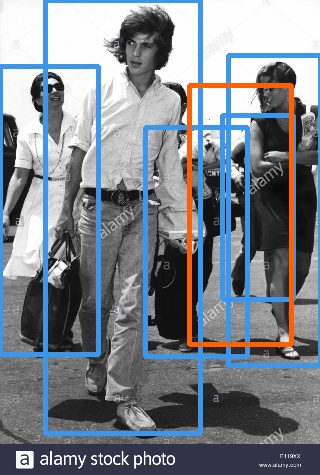} &
  \includegraphics[width=0.15\textwidth,trim={0 1.1cm 0 0},clip]{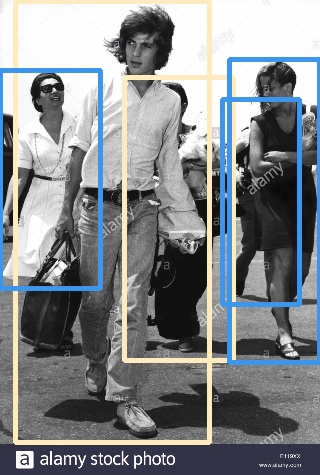} &
  \includegraphics[width=0.15\textwidth,trim={0 1.1cm 0 0},clip]{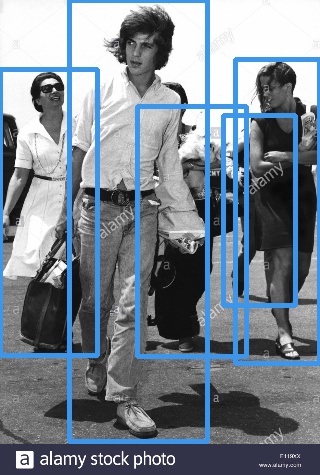} &
  \includegraphics[width=0.15\textwidth,trim={0 1.1cm 0 0},clip]{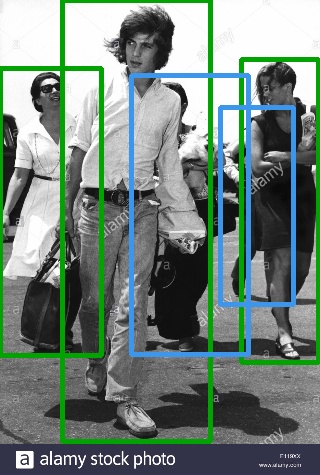} &
  \includegraphics[width=0.15\textwidth,trim={0 1.1cm 0 0},clip]{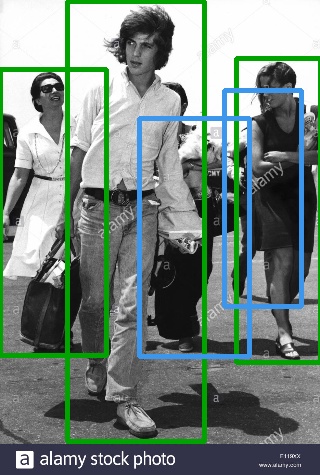} \\

  \includegraphics[width=0.15\textwidth]{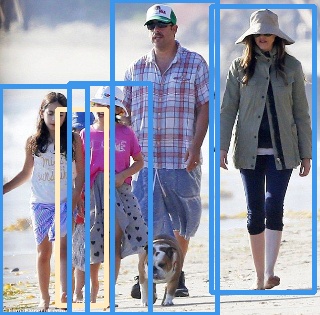} &
  \includegraphics[width=0.15\textwidth]{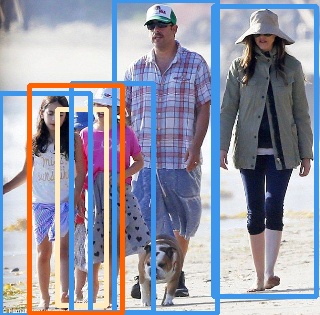} &
  \includegraphics[width=0.15\textwidth]{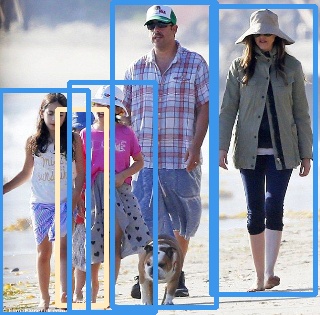} &
  \includegraphics[width=0.15\textwidth]{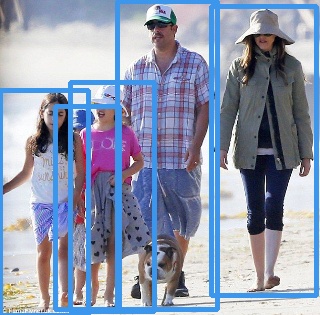} &
  \includegraphics[width=0.15\textwidth]{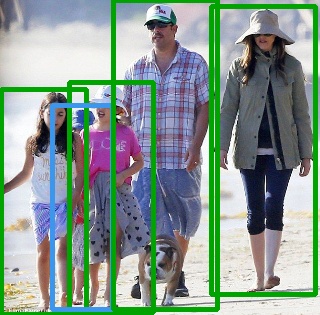} &
  \includegraphics[width=0.15\textwidth]{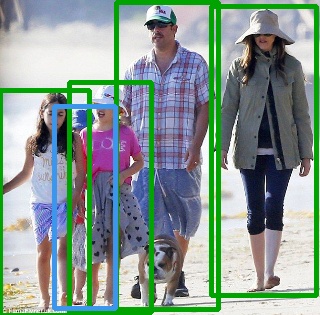} \\

  \includegraphics[width=0.15\textwidth]{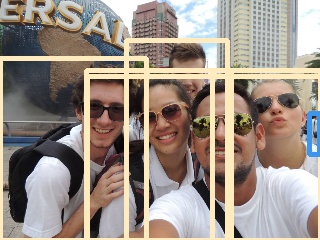} &
  \includegraphics[width=0.15\textwidth]{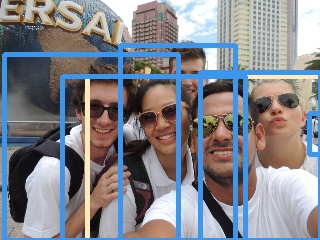} &
  \includegraphics[width=0.15\textwidth]{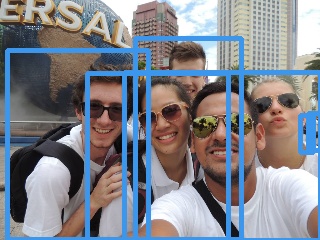} &
  \includegraphics[width=0.15\textwidth]{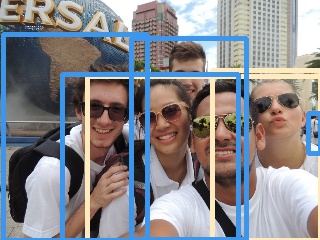} &
  \includegraphics[width=0.15\textwidth]{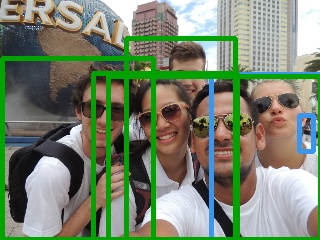} &
  \includegraphics[width=0.15\textwidth]{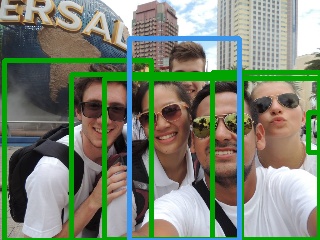} \\

  \includegraphics[width=0.15\textwidth]{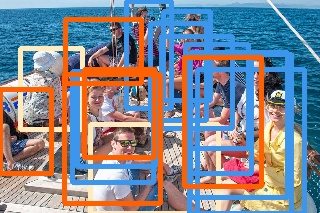} &
  \includegraphics[width=0.15\textwidth]{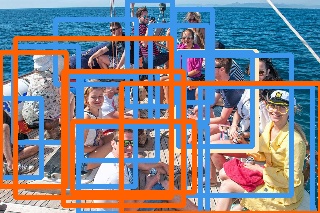} &
  \includegraphics[width=0.15\textwidth]{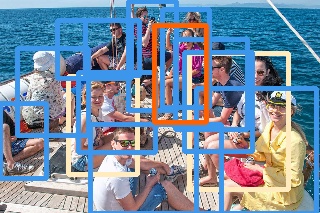} &
  \includegraphics[width=0.15\textwidth]{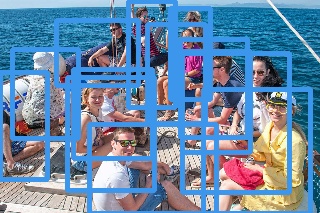} &
  \includegraphics[width=0.15\textwidth]{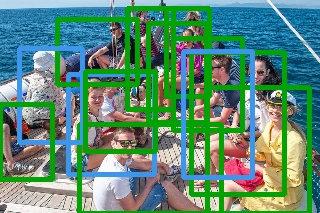} &
  \includegraphics[width=0.15\textwidth]{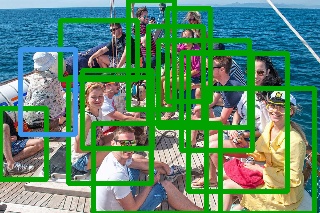} \\

  \includegraphics[width=0.15\textwidth]{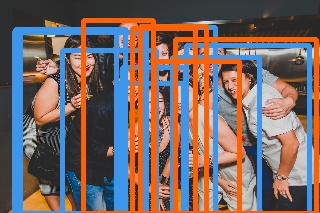} &
  \includegraphics[width=0.15\textwidth]{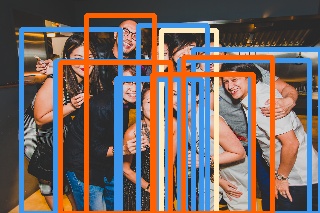} &
  \includegraphics[width=0.15\textwidth]{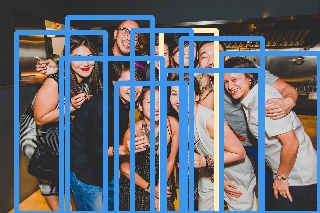} &
  \includegraphics[width=0.15\textwidth]{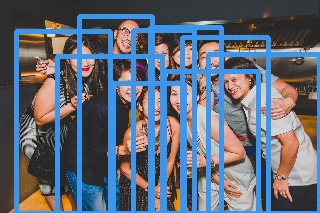} &
  \includegraphics[width=0.15\textwidth]{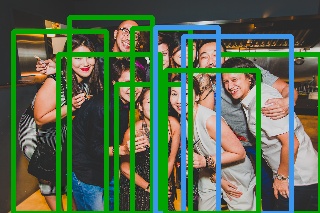} &
  \includegraphics[width=0.15\textwidth]{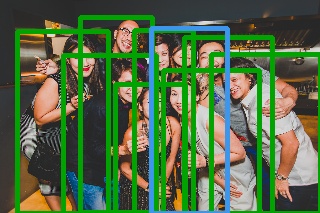} \\

  \includegraphics[width=0.15\textwidth,trim={0 0 0 .8cm},clip]{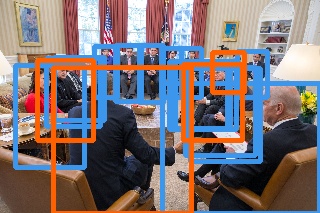} &
  \includegraphics[width=0.15\textwidth,trim={0 0 0 .8cm},clip]{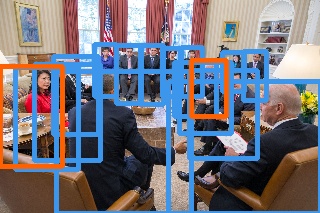} &
  \includegraphics[width=0.15\textwidth,trim={0 0 0 .8cm},clip]{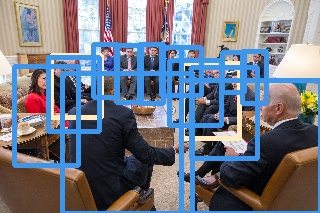} &
  \includegraphics[width=0.15\textwidth,trim={0 0 0 .8cm},clip]{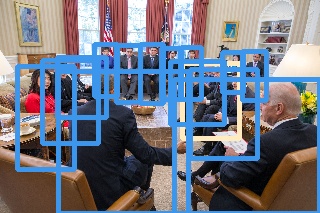} &
  \includegraphics[width=0.15\textwidth,trim={0 0 0 .8cm},clip]{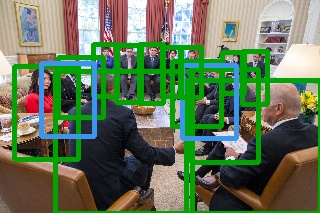} &
  \includegraphics[width=0.15\textwidth,trim={0 0 0 .8cm},clip]{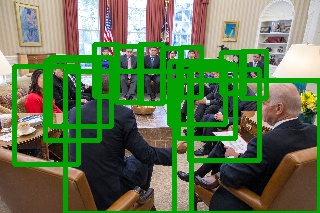} \\

  \includegraphics[width=0.15\textwidth]{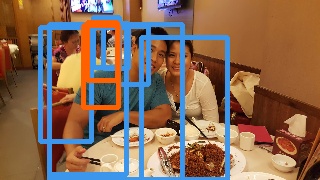} &
  \includegraphics[width=0.15\textwidth]{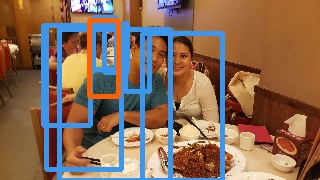} &
  \includegraphics[width=0.15\textwidth]{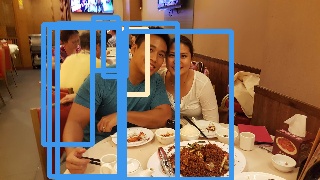} &
  \includegraphics[width=0.15\textwidth]{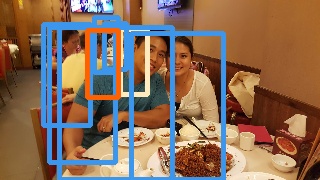} &
  \includegraphics[width=0.15\textwidth]{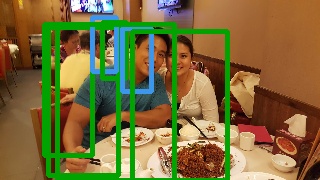} &
  \includegraphics[width=0.15\textwidth]{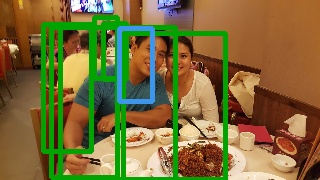} \\

\bottomrule
\end{tabular}

  \caption{Results visualization of RelationNet~\cite{hu2018relation}, IterDet~\cite{iterdet2021}, Sparse RCNN~\cite{sun2020sparse}, deformable DETR~\cite{zhu2021deformable} and our approach based on them~\cite{sun2020sparse, zhu2021deformable}. Blue boxes are true positive detections, light yellow boxes are missed instances and orange boxes are false positives. Green boxes represent progressively refined detections in our method.}
\end{figure*}
\clearpage
\end{document}


\appendix

\section{Implementation of deformable DETR~\cite{zhu2021deformable} with progressive predicting strategy.}

We also equip the deformable DETR~\cite{zhu2021deformable} with our progressive predicting schema to demonstrate the generality of our method. The decoder in Deformable DETR consists of 6 stages. The architecture of decoding stages are depicted in Figure.~\ref{fig:ddetr_arch}. As describe in the manuscript, we plug our designed components into the last decoder stage, which are identical to that embeded into Sparse RCNN~\cite{sun2020sparse}. It is also depicted in Figure.~\ref{fig:sr_ddetr}, the hyper-parameters, e.g.\ the feature dimension in relation information extractor $\mathcal{R}$, are also identical to those adopted in Sparse RCNN.

We choose the deformable DETR with iterative bounding box refinement. Following deformable DETR~\cite{zhu2021deformable}, we use \emph{ResNet-50} that pre-trained on \emph{ImageNet}~\cite{imagenet} as backbone. The model is trained with Adam optimizer with a weight decay of 0.0001. The total training duration is 50 epochs on 8 GPUs with 1 image per GPU. The initial learning rate is 0.0002 and dropped by a factor of 0.1 after 40 epochs. The parameters initialization in the newly added components and losses weights are identical to the original work~\cite{zhu2021deformable}. The default number of queries and stages is 500 and 6, respectively. The hyper-parameters $s$ and $\theta$ are also $0.7$ and $0.4$. The gradients are detached at proposal boxes from the second stage to stabilize training. We stop gradient back-propagation from the last stage to its previous one. Additionally, those negative samples who overlaps with ignore region with an \textit{intersection-over-area}(ioa) greater than $0.7$ are not involved in training.



\begin{figure}[!t]
  \centering
  \begin{subfigure}{\linewidth}
    \definecolor{brown}{HTML}{843C0C}
\definecolor{darkred}{HTML}{C43C0C}
\definecolor{skyblue}{HTML}{00B0F0}
\definecolor{black}{HTML}{000000}

\begin{tikzpicture}[
    scale=0.9,
    sym/.style={inner sep=1},
    box/.style={rectangle, rounded corners=3, draw=black, thick, minimum size=16, minimum width=54},
    roi/.style={rectangle, rounded corners=3, draw=black, thick, minimum size=16, inner sep=0, fill=white},
    rnet/.style={rectangle, minimum size=16},
    dr/.style={rectangle, rounded corners=3, draw=black, thick, minimum height=16, inner sep=1, fill=white},
    arr/.style={thick, ->, >=stealth, draw=black}
  ]
  \small

  \node[sym]   at (0, 0)       (prev_q)              {$\mathbf{q}_{t-1}$};
  \node[sym]   at (0, 3)       (prev_box)            {$\mathbf{b}_{t-1}$};
  \node[box]   at (2, 1.5)     (dyn_conv)            {MCA$_{t-1}$};
  \node[box]   at (2, 0)       (self_attn)           {MSA$_{t-1}$};
  \node[sym]   at (2, 3)       (box)                 {$\mathbf{b}_{t}$};
  \node[sym]   at (.2+4, 0)    (q)                   {$\mathbf{q}_{t}$};
  \node[box]   at +(.4+6, 1.5) (_dyn_conv)           {MCA$_t$};
  \node[box]   at +(.4+6, 0)   (_self_attn)          {MSA$_t$};
  \node[sym]   at +(.4+8, 0)   (_q)                  {$\mathbf{q}_{t+1}$};
  \node[sym]   at +(.4+6, 3)   (_box)                {$\mathbf{b}_{t+1}$};

  \draw[arr] (prev_box)         -- (dyn_conv.west);
  \draw[arr] (prev_q)           -- (self_attn);
  \draw[arr] (self_attn)        -- (dyn_conv);
  \draw[arr] (dyn_conv)         -- (box);
  \draw[arr] (dyn_conv.east)    -- (q);
  \draw[arr] (box) -- (.2+4, 3) -- (_dyn_conv.west);
  \draw[arr] (_self_attn)       -- (_dyn_conv);
  \draw[arr] (_dyn_conv)        -- (_box);
  \draw[arr] (_dyn_conv.east)   -- (_q);
  \draw[arr] (q)                -- (_self_attn);
\end{tikzpicture}
    \caption{Decoder in deformable DETR~\cite{zhu2021deformable}. $\text{MCA}$ -- multi-head cross-attention, $\text{MSA}$ -- multi-head self-attention.}
    \label{fig:ddetr_arch}
  \end{subfigure}
  \vspace{-0.5pc}
  \begin{subfigure}{\linewidth}
    \definecolor{brown}{HTML}{843C0C}
\definecolor{darkred}{HTML}{C43C0C}
\definecolor{skyblue}{HTML}{00B0F0}
\definecolor{black}{HTML}{000000}
\begin{tikzpicture}[
    scale=0.9,
    sym/.style={inner sep=1},
    box/.style={rectangle, rounded corners=3, draw=black, thick, minimum size=16, minimum width=54},
    roi/.style={rectangle, rounded corners=3, draw=black, thick, minimum size=16, inner sep=0, fill=white},
    rnet/.style={rectangle, minimum size=16},
    dr/.style={rectangle, rounded corners=3, draw=black, thick, minimum height=16, inner sep=1, fill=white},
    arr/.style={thick, ->, >=stealth, draw=black}
  ]
  \small
  \node[sym]             at (0, 0)       (prev_q)        {$\mathbf{q}_{t-1}$};
  \node[sym]             at (0, 3)       (prev_box)      {$\mathbf{b}_{t-1}$};
  \node[box]             at (2, 1.5)     (dyn_conv)      {MCA$_{t-1}$};
  \node[box]             at (2, 0)       (self_attn)     {MSA$_{t-1}$};
  \node[sym]             at (2, 3)       (box)           {$\mathbf{b}_{t}$};
  \node[box, draw=none]  at (2, 3)       (hidden_box)    {};
  \node[sym]             at (.2+4, 0)    (q)             {$\mathbf{q}_{t}$};
  \node[box]             at +(.4+6, 1.5) (_dyn_conv)     {MCA$_t$};
  \node[box]             at +(.4+6, 0)   (_self_attn)    {LMSA$_t$};
  \node[sym]             at +(.4+8, 0)   (_q)            {$\mathbf{q}_{t+1}$};
  \node[sym]             at +(.4+6, 3)   (_box)          {$\mathbf{b}_{t+1}$};
  \node[dr, anchor=east] at (.2+4+.01, 1.5)  (d)             {$\mathcal{S}$};
  \node[dr, anchor=west] at (.2+4-.01, 1.5)  (r)             {$\mathcal{R}$};
  \node[rnet]            at (.2+4, 1.5)  (rnet)          {};

  \draw[arr] (prev_box)         -- (dyn_conv.west);
  \draw[arr] (prev_q)           -- (self_attn);
  \draw[arr] (self_attn)        -- (dyn_conv);
  \draw[arr] (dyn_conv)         -- (box);
  \draw[arr] (dyn_conv.east)    -- (q);
  \draw[arr] (box) -- (.2+4, 3) -- (_dyn_conv.west);
  \draw[arr] (hidden_box.east)  -- (rnet);
  \draw[arr] (rnet)             -- (_self_attn.west);
  \draw[arr] (_self_attn)       -- (_dyn_conv);
  \draw[arr] (_dyn_conv)        -- (_box);
  \draw[arr] (_dyn_conv.east)   -- (_q);
  \draw[arr] (q)                -- (.2+4, 1.2);
\end{tikzpicture}
    \caption{Decoder in SR-Deformable DETR (Ours). $\mathcal{S}$ -- Prediction Selector, $\mathcal{R}$ -- Relation information extractor, $\text{LMSA}$ -- local multi-head self-attention.}
    \label{fig:sr_ddetr}
  \end{subfigure}
  \vspace{-0.5pc}
  \caption{~\ref{fig:ddetr_arch} is the architecture of decoding stage in deformable DETR~\cite{zhu2021deformable};
  ~\ref{fig:sr_ddetr} describes the decoding stage structure equipped with our designed components for progressive predicting schema. }
  
\end{figure}

\section{Another new label assignment strategy.}

\begin{algorithm}[ht] 
\caption{Label Assignment for $\mathcal{D}^l_{t}$.} 
\label{alg:new_label_assign} 
\begin{algorithmic}[1] 
\Require 
$\mathcal{D}^{l}_{t}$, $ \mathcal{D}^{h}_{t}$, $\mathcal{G}$; \\
 $\mathcal{D}^{l}_{t}$: results of $\mathcal{D}^{l}_{t-1}$ in Equ.(1) after stage $t$; \\
 $\mathcal{D}^{h}_{t}$:  results of $\mathcal{D}^{h}_{t-1}$ in Equ.(1) after stage $t$;\\
$\mathcal{G}$: target boxes.
\Ensure 
The target boxes assignment for $\mathcal{D}^{l}_{t}$.
\State Merge $\mathcal{D}^{h}_{t}$ and $\mathcal{D}^{l}_{t}$: \quad
$\mathcal{D}_{t} = \mathcal{D}^{h}_{t} \bigcup \mathcal{D}^{l}_{t}$;
\State Compute matching costs $\mathcal{C}_{t}$ between $\mathcal{D}_{t}$ and $\mathcal{G}$;
\State $\mathcal{M}_G, \mathcal{M}_{D} = \text{HungarianMatch}(\mathcal{D}_{t}, \mathcal{G}, \mathcal{C}_{t})$;
\State Acquire the matched predictions $\mathcal{M}^h_{D}$ and $\mathcal{M}^h_{\mathcal{G}}$: \\
$\mathcal{M}^h_{D}=\{ b|<b,g>, b \in \mathcal{D}^h_t \wedge g \in \mathcal{M}_{G}\}$. \\
$\mathcal{M}^h_{\mathcal{G}}=\{g| <b,g>, b \in \mathcal{D}^h_{t} \wedge g \in \mathcal{M}_{G} \}$
\State Differentiate $\mathcal{M}^{l}_G, \mathcal{M}^{l}_{D}$ from $\mathcal{M}^h_{D}$, $\mathcal{M}^h_{G}$ in $\mathcal{M}_G, \mathcal{M}_{D}$: \\
$\mathcal{M}^{l}_G = \mathcal{M}_{G}- \mathcal{M}^h_{G};$\\
$\mathcal{M}^{l}_{D_{t}}= \mathcal{M}_{D} - \mathcal{M}^h_{D}$; \\
\Return $\mathcal{M}^{l}_G, \mathcal{M}^{l}_{D}$; 
\end{algorithmic}\label{algorithm:first}
\end{algorithm}

After the accomplishment of the experiments with \emph{query-based} detector, .e.g.\ deformable DETR~\cite{zhu2021deformable} and Sparse RCNN~\cite{sun2020sparse}, we design a new label assignment for training the last decoder stage, which is described in Algorithm.~\ref{alg:new_label_assign}. All the symbols share the same meaning as that in the original manuscript. e.g.\ $<b, g>$ is a matched pair after performing Hungarian Matching. b and g denote a predicted box and its target box, respectively. $\mathcal{M}^{h}_{D}$ are the boxes from accepted predictions while $\mathcal{M}^h_{G}$ represents their matched target boxes. Table.~\ref{tbl:label_assign} describes the performance comparisons with other methods. The new label assignment is comparable to that described in the manuscript when evaluated on the \emph{CrowdHuman}~\cite{shao2018crowdhuman} dataset with that deployed to Sparse RCNN~\cite{sun2020sparse} while a slightly worse in $\text{MR}^{-2}$ in deformable DETR~\cite{zhu2021deformable}.

\begin{table}[ht]
	\centering
	\begin{tabular}{llccc}
		\toprule
		Method & \#Queries & AP & $\text{MR}^{-2}$  & JI \\
		\hline
		S-RCNN~\cite{sun2020sparse} & 500 & 90.7 & 44.7 & 81.4 \\
		D-DETR \cite{zhu2021deformable} & 1000 & 91.5 & 43.7 & 83.1 \\
		\hline
		S-RCNN+\emph{Ours} & 500 & 92.0 & 41.4 & 83.2 \\
		D-DETR+\emph{Ours} & 1000 & 92.1 & 41.5 & 84.0 \\
		\hline
		$\text{S-RCNN+\emph{Ours}}^{\ddag}$ & 500 & 92.1 & 41.6 & 83.2 \\
		$\text{D-DETR+\emph{Ours}}^{\ddag}$ & 1000 & 92.1 & 42.5 & 84.2 \\
		\bottomrule
	\end{tabular}
	\caption{Comparisons of different label assignment on \emph{CrowdHuman} validation set.$\text{}^{\ddag}$ means utilizing the newly label assignment to train a \textit{query-based} detector.}
	\label{tbl:label_assign}
	\vspace{-1.5pc}
\end{table}

\begin{figure*}
  \small
  \centering
  \setlength{\tabcolsep}{3pt}
\begin{tabular}{cccccc}
  \toprule
  RelationNet~\cite{hu2018relation} & IterDet~\cite{iterdet2021} & Sparse R-CNN~\cite{sun2020sparse} & D-DETR~\cite{carion2020end} & Sparse RCNN+Ours & D-DETR+Ours \\
  \midrule

  \includegraphics[width=0.15\textwidth,trim={0 0 0 2.4cm},clip]{images/figs/273278,12d5a00000b7cb45_relation.jpg} &
  \includegraphics[width=0.15\textwidth,trim={0 0 0 2.4cm},clip]{images/figs/273278,12d5a00000b7cb45_iterdet.jpg} &
  \includegraphics[width=0.15\textwidth,trim={0 0 0 2.4cm},clip]{images/figs/273278,12d5a00000b7cb45_sparse-rcnn-new.jpg} &
  \includegraphics[width=0.15\textwidth,trim={0 0 0 2.4cm},clip]{images/figs/273278,12d5a00000b7cb45_detr.jpg} &
  \includegraphics[width=0.15\textwidth,trim={0 0 0 2.4cm},clip]{images/figs/273278,12d5a00000b7cb45_iter-sparse-rcnn.jpg} &
  \includegraphics[width=0.15\textwidth,trim={0 0 0 2.4cm},clip]{images/figs/273278,12d5a00000b7cb45_iter.detr.jpg} \\

  \includegraphics[width=0.15\textwidth]{images/figs/282555,16a5a000e5588af5_relation.jpg} &
  \includegraphics[width=0.15\textwidth]{images/figs/282555,16a5a000e5588af5_iterdet.jpg} &
  \includegraphics[width=0.15\textwidth]{images/figs/282555,16a5a000e5588af5_sparse-rcnn-new.jpg} &
  \includegraphics[width=0.15\textwidth]{images/figs/282555,16a5a000e5588af5_detr.jpg} &
  \includegraphics[width=0.15\textwidth]{images/figs/282555,16a5a000e5588af5_iter-sparse-rcnn.jpg} &
  \includegraphics[width=0.15\textwidth]{images/figs/282555,16a5a000e5588af5_iter.detr.jpg} \\
  
  \includegraphics[width=0.15\textwidth,trim={0 1.1cm 0 0},clip]{images/figs/283991,1f5900026d74f0d_relation.jpg} &
  \includegraphics[width=0.15\textwidth,trim={0 1.1cm 0 0},clip]{images/figs/283991,1f5900026d74f0d_iterdet.jpg} &
  \includegraphics[width=0.15\textwidth,trim={0 1.1cm 0 0},clip]{images/figs/283991,1f5900026d74f0d_sparse-rcnn-new.jpg} &
  \includegraphics[width=0.15\textwidth,trim={0 1.1cm 0 0},clip]{images/figs/283991,1f5900026d74f0d_detr.jpg} &
  \includegraphics[width=0.15\textwidth,trim={0 1.1cm 0 0},clip]{images/figs/283991,1f5900026d74f0d_iter-sparse-rcnn.jpg} &
  \includegraphics[width=0.15\textwidth,trim={0 1.1cm 0 0},clip]{images/figs/283991,1f5900026d74f0d_iter.detr.jpg} \\

  \includegraphics[width=0.15\textwidth]{images/figs/283554,2605000054a40402_relation.jpg} &
  \includegraphics[width=0.15\textwidth]{images/figs/283554,2605000054a40402_iterdet.jpg} &
  \includegraphics[width=0.15\textwidth]{images/figs/283554,2605000054a40402_sparse-rcnn-new.jpg} &
  \includegraphics[width=0.15\textwidth]{images/figs/283554,2605000054a40402_detr.jpg} &
  \includegraphics[width=0.15\textwidth]{images/figs/283554,2605000054a40402_iter-sparse-rcnn.jpg} &
  \includegraphics[width=0.15\textwidth]{images/figs/283554,2605000054a40402_iter.detr.jpg} \\

  \includegraphics[width=0.15\textwidth]{images/figs/282555,baa5a00000be1da2_relation.jpg} &
  \includegraphics[width=0.15\textwidth]{images/figs/282555,baa5a00000be1da2_iterdet.jpg} &
  \includegraphics[width=0.15\textwidth]{images/figs/282555,baa5a00000be1da2_sparse-rcnn-new.jpg} &
  \includegraphics[width=0.15\textwidth]{images/figs/282555,baa5a00000be1da2_detr.jpg} &
  \includegraphics[width=0.15\textwidth]{images/figs/282555,baa5a00000be1da2_iter-sparse-rcnn.jpg} &
  \includegraphics[width=0.15\textwidth]{images/figs/282555,baa5a00000be1da2_iter.detr.jpg} \\

  \includegraphics[width=0.15\textwidth]{images/figs/273271,2277e000891f2e71_relation.jpg} &
  \includegraphics[width=0.15\textwidth]{images/figs/273271,2277e000891f2e71_iterdet.jpg} &
  \includegraphics[width=0.15\textwidth]{images/figs/273271,2277e000891f2e71_sparse-rcnn-new.jpg} &
  \includegraphics[width=0.15\textwidth]{images/figs/273271,2277e000891f2e71_detr.jpg} &
  \includegraphics[width=0.15\textwidth]{images/figs/273271,2277e000891f2e71_iter-sparse-rcnn.jpg} &
  \includegraphics[width=0.15\textwidth]{images/figs/273271,2277e000891f2e71_iter.detr.jpg} \\

  \includegraphics[width=0.15\textwidth]{images/figs/273271,1ddda0001685370d_relation.jpg} &
  \includegraphics[width=0.15\textwidth]{images/figs/273271,1ddda0001685370d_iterdet.jpg} &
  \includegraphics[width=0.15\textwidth]{images/figs/273271,1ddda0001685370d_sparse-rcnn-new.jpg} &
  \includegraphics[width=0.15\textwidth]{images/figs/273271,1ddda0001685370d_detr.jpg} &
  \includegraphics[width=0.15\textwidth]{images/figs/273271,1ddda0001685370d_iter-sparse-rcnn.jpg} &
  \includegraphics[width=0.15\textwidth]{images/figs/273271,1ddda0001685370d_iter.detr.jpg} \\

  \includegraphics[width=0.15\textwidth,trim={0 0 0 .8cm},clip]{images/figs/273275,10cd9200055f04d4a_relation.jpg} &
  \includegraphics[width=0.15\textwidth,trim={0 0 0 .8cm},clip]{images/figs/273275,10cd9200055f04d4a_iterdet.jpg} &
  \includegraphics[width=0.15\textwidth,trim={0 0 0 .8cm},clip]{images/figs/273275,10cd9200055f04d4a_sparse-rcnn-new.jpg} &
  \includegraphics[width=0.15\textwidth,trim={0 0 0 .8cm},clip]{images/figs/273275,10cd9200055f04d4a_detr.jpg} &
  \includegraphics[width=0.15\textwidth,trim={0 0 0 .8cm},clip]{images/figs/273275,10cd9200055f04d4a_iter-sparse-rcnn.jpg} &
  \includegraphics[width=0.15\textwidth,trim={0 0 0 .8cm},clip]{images/figs/273275,10cd9200055f04d4a_iter.detr.jpg} \\

  \includegraphics[width=0.15\textwidth]{images/figs/273278,103204000dfd1be77_relation.jpg} &
  \includegraphics[width=0.15\textwidth]{images/figs/273278,103204000dfd1be77_iterdet.jpg} &
  \includegraphics[width=0.15\textwidth]{images/figs/273278,103204000dfd1be77_sparse-rcnn-new.jpg} &
  \includegraphics[width=0.15\textwidth]{images/figs/273278,103204000dfd1be77_detr.jpg} &
  \includegraphics[width=0.15\textwidth]{images/figs/273278,103204000dfd1be77_iter-sparse-rcnn.jpg} &
  \includegraphics[width=0.15\textwidth]{images/figs/273278,103204000dfd1be77_iter.detr.jpg} \\

\bottomrule
\end{tabular}

  \caption{Results visualization of RelationNet~\cite{hu2018relation}, IterDet~\cite{iterdet2021}, Sparse RCNN~\cite{sun2020sparse}, deformable DETR~\cite{zhu2021deformable} and our approach based on them~\cite{sun2020sparse, zhu2021deformable}. Blue boxes are true positive detections, light yellow boxes are missed instances and orange boxes are false positives. Green boxes represent progressively refined detections in our method.}
\end{figure*}

\section{Changes in Performance According to Depth.}
The performance of a \emph{query-based} detector would not be improved but will degrade as the depth of a decoder increases when handling crowded scenes. We conduct an experiment on the \emph{CrowdHuman} dataset, taking Sparse RCNN with \emph{ResNet-50} as the base detector and equip it with 500 queries. We adjust the depth of its decoder while remaining the others unchanged. As is described in Table. ~\ref{tbl:depth_exp}, the performance degrades as the depth of the decoder increases.

\begin{table}[ht]
	\centering
	\begin{tabular}{lccc}
		\toprule
		 \#Depth & AP & $\text{MR}^{-2}$  & JI \\
		\hline
		 6 & 90.7 & 44.7 & 81.4 \\
		 7 & 90.6 & 45.7 & 81.0 \\
		8 & 90.4 & 45.9 & 80.3 \\
		9 & 90.7 & 44.4 & 80.9 \\
        10 & 90.2 & 46.6 & 80.0 \\
		\bottomrule
	\end{tabular}
	\caption{Experiment analysis as the depth of a decoder increases, which performs on \emph{CrowdHuman} dataset.}
	\label{tbl:depth_exp}
	\vspace{-1.5pc}
\end{table}


{\small
\bibliographystyle{ieee_fullname}
\bibliography{ref21, ref22}
}